\documentclass[10pt,twocolumn,letterpaper]{article}

\usepackage[english]{babel}
\usepackage{algorithm}
\usepackage{algorithmic}
\usepackage{abstract}
\usepackage{graphicx}
\usepackage{color}
\usepackage{transparent}
\usepackage{verbatim}
\usepackage{amsmath}
\usepackage{capt-of,etoolbox}
\usepackage[caption=false]{subfig}
\usepackage{paralist}
\usepackage{amsthm}
\usepackage{enumitem}
\usepackage{amsfonts}
\usepackage{iccv}
\usepackage{times}
\usepackage{epsfig}
\usepackage{amssymb}
\usepackage[autostyle, english = american]{csquotes}

\usepackage[bookmarks=false]{hyperref}

\usepackage{array}
\newcommand{\head}[2]{\multicolumn{1}{>{\centering\arraybackslash}p{#1}|}{\textbf{#2}}}

\newcommand{\IC}[1]{I^{#1}}
\newcommand{\ID}[1]{I_{#1}}

\newcommand{\CAM}{C}

\newcommand{\setD}{\mathcal{I}}

\newcommand{\setP}{\mathcal{P}}

\MakeOuterQuote{"}

\DeclareMathOperator*{\argmin}{arg\,min}

\iccvfinalcopy

\theoremstyle{definition}

\hypersetup{
    bookmarks=false,         
    unicode=true,          
    pdftoolbar=true,        
    pdfmenubar=true,        
    pdffitwindow=false,     
    pdfstartview={FitH},    
    pdftitle={My title},    
    pdfauthor={Author},     
    pdfsubject={Subject},   
    pdfcreator={Creator},   
    pdfproducer={Producer}, 
    pdfkeywords={keyword1} {key2} {key3}, 
    pdfnewwindow=true,      
    colorlinks=true,       
    linkcolor=magenta,          
    citecolor=blue,        
    filecolor=red,      
    urlcolor=cyan,          
		pdfpagemode=UseNone
}

\begin{document}

\title{Have a Look at What I See}
\author{Lior Talker\\
  The University of Haifa, Israel\\
  ltalke01@campus.haifa.ac.il\\
	talker.lior@idc.ac.il
\and Yael Moses \\
  The Interdisciplinary Center, Israel\\
  yael@idc.ac.il
	\and Ilan Shimshoni\\
  The University of Haifa, Israel\\
  ishimshoni@mis.haifa.ac.il}

\renewcommand{\tabcolsep}{1pt}
\makeatletter
\let\@oldmaketitle\@maketitle
\renewcommand{\@maketitle}{\@oldmaketitle%
\begin{tabular}{@{}ccccc@{}}
  \includegraphics[width=0.195\linewidth]{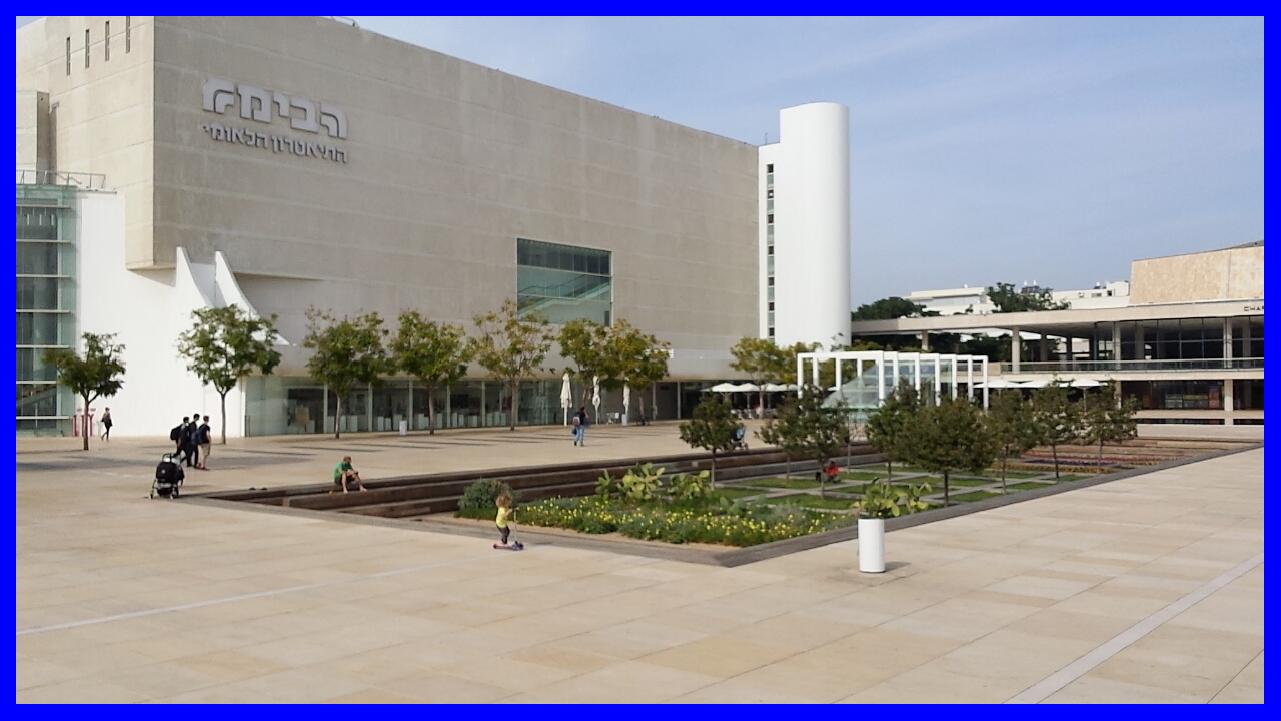} &
	\includegraphics[width=0.195\linewidth]{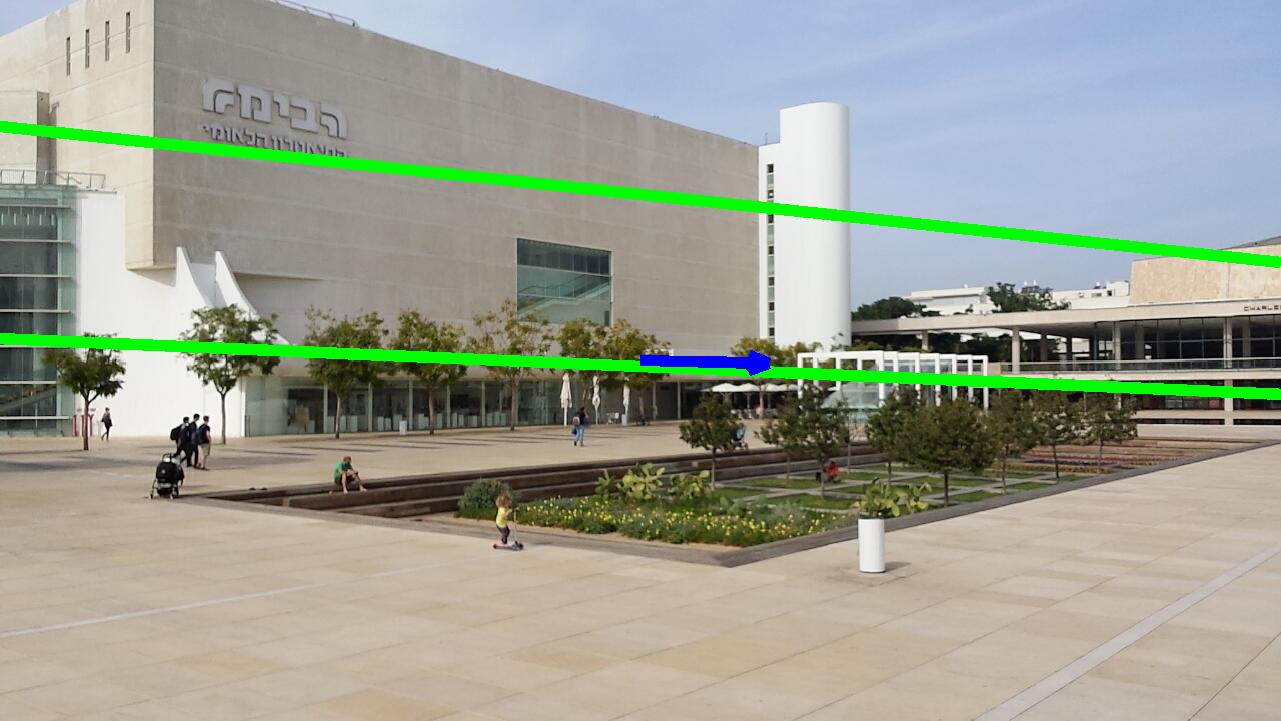}  &
	\includegraphics[width=0.195\linewidth]{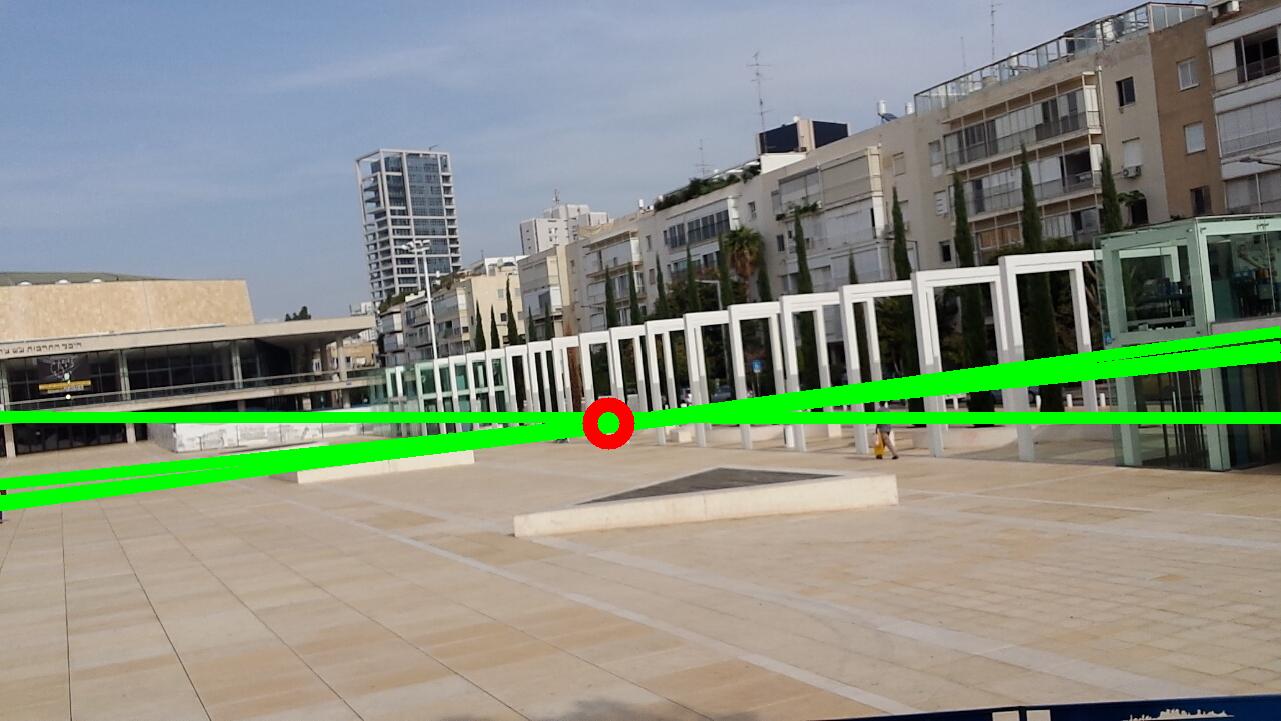} &
	\includegraphics[width=0.195\linewidth]{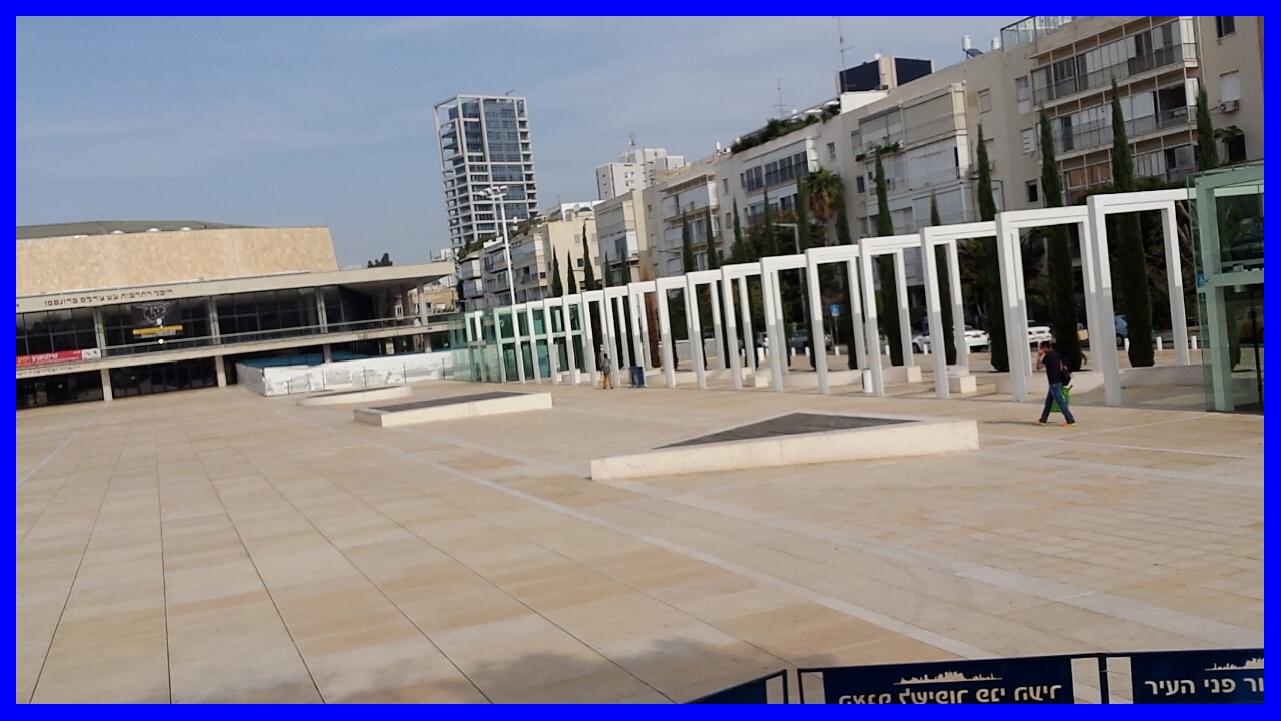} &
	\includegraphics[width=0.195\linewidth]{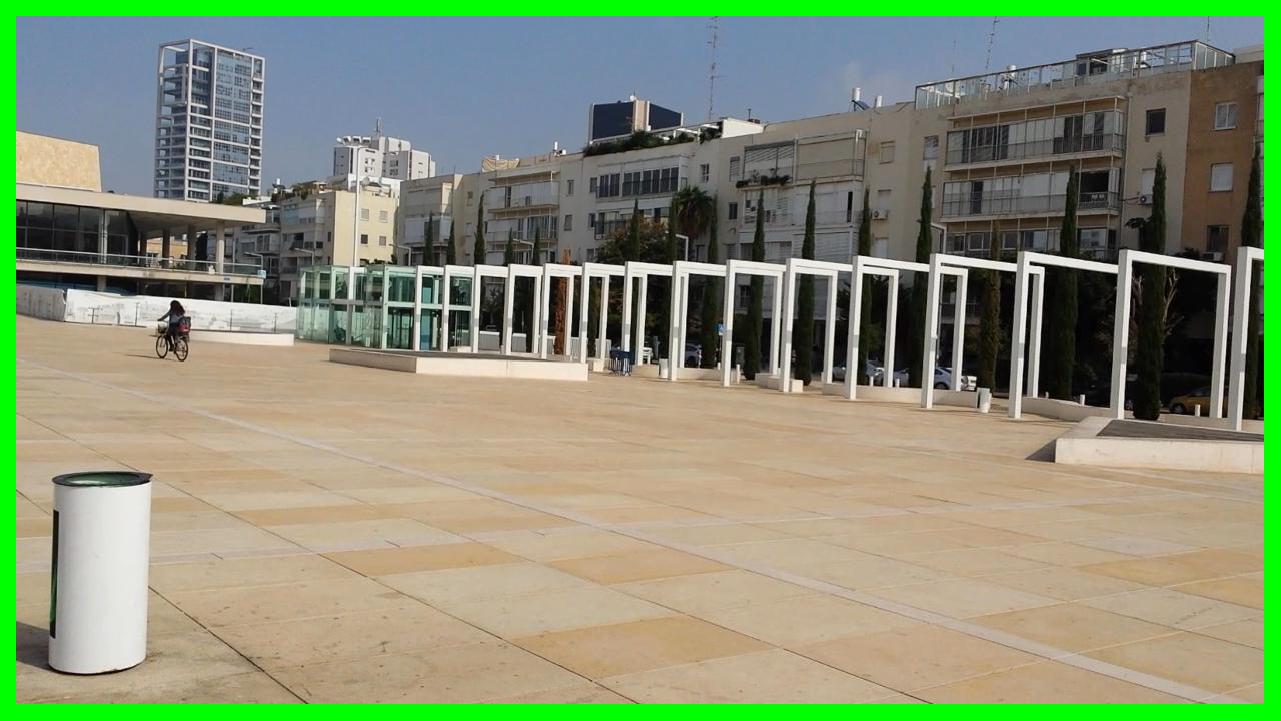} \\
Initial & Initial + interface & First iter.~ + interface & Final & Destination
\end{tabular}
	\bigskip}
\makeatother
\maketitle

\begin{abstract}
We propose a method for guiding a photographer to rotate her/his smartphone camera to obtain an image that overlaps with another image of the same scene. The other image is taken by another photographer from a different viewpoint. Our method is applicable even when the images do not have overlapping fields of view. 
Straightforward applications of our method include sharing attention to regions of interest for social purposes, or adding missing images to improve structure for motion results.
Our solution uses additional images of the scene, which are often available since  many people use their smartphone cameras regularly. These images may be available online from other photographers who are present at the scene.
Our method avoids 3D scene reconstruction; it relies instead on a new representation that consists of the spatial orders of the scene points on two axes, $x$ and $y$.
This representation allows a sequence of points to be chosen efficiently and projected onto the photographer's images, using epipolar point transfer. Overlaying these epipolar lines on the live preview of the camera produces a convenient interface to guide the user. The method was tested on challenging datasets of images and succeeded in guiding a photographer from one view to a non-overlapping destination view.


\end{abstract}

\section{Introduction}

Assume Alice and Bob capture two images of non-overlapping sections of the same scene. Alice then says to Bob, ``Have a look at what I see.'' Can Bob rotate his camera to view the section of the scene viewed by Alice?
In this paper we refer to the problem of computing this desired rotation and conveying it to the photographer as the {\em camera guidance problem}. 




The objective in the camera guidance problem is to guide Bob's camera to rotate and capture a new image that significantly overlaps with Alice's image (i.e., the images share pixels that correspond to the same scene points). To simplify the objective, we aim to determine the rotation of Bob's camera such that a scene point, $P_{\scriptscriptstyle T}$, projected to the center of Alice's image (the destination image) will also be projected to Bob's newly captured image.


We propose a method to compute the projection of the scene point, $P_{\scriptscriptstyle T}$, to the initial image captured by Bob (the {\em initial image}).
We then show how this point can be used to compute the desired rotation and convey it to the user via a simple interface.
Our solution makes use of additional images of the scene that are assumed to be available from other photographers that are present in the scene. Such additional images are necessary to solve the problem when the destination and initial images do not overlap. 

Before we describe our method, let us first consider two existing methods that are natural candidates to compute the projection of $P_{\scriptscriptstyle T}$ to the initial image's plane. As we will show in Section \ref{sec:method} and demostrated in Section \ref{sec:experiments}, Structure from Motion (SFM) methods \cite{szeliski2010computer, forsyth2002computer, wu2013towards, crandall2011discrete, sinha2012multi} and image-based panorama \cite{brown2007automatic,szeliski2010computer}, two seemingly natural candidates for computing the projection of $P_{\scriptscriptstyle T}$ to the initial image's plane, are not effective in the scenarios we consider. The former is time-consuming and requires a large number of images and the latter is not applicable to general 3D scenes such as the ones we consider here.

We propose instead to efficiently choose a small partial set of images and use epipolar constraints to compute the projection of $P_{\scriptscriptstyle T}$
 to the initial image plane. When the gap between the initial and the destination images is large, the goal is achieved through a sequence of intermediate views. Each such view corresponds to a rotation of the camera such that a projection of a chosen scene point is at the center region of the new captured image.


To efficiently choose the partial set of images, we propose a novel rough representation of the scene. The representation consists of the spatial orders of the scene points on two axes, $x$ and $y$. We call this representation \emph{Spatially Ordered Feature Aggregation (SOFA)}. Computing SOFA is trivial if reliable feature correspondence is available and the spatial orders of the scene points 
are preserved in their projections to the images. In this case, the axes of a single image can be used to represent the order of the features as all partial orders are consistent. However, all existing feature matching methods are imperfect, and in practice, for most scenes, the order-preserving assumption does not hold.
Hence we propose to combine the orderings of the features in the different views to obtain approximate global orderings. To achieve this goal in a robust manner we use the rank aggregation method~\cite{dwork2001rank}. Rank aggregation methods were employed for combining the ranks of web pages obtained by several search engines and were previously used in computer vision for ranking the temporal order of images~\cite{dekel2014photo}.


Finally, we need to convey to the photographer how to rotate the camera towards the goal. When the camera is automatically controlled (e.g., a camera mounted on a robot or a PTZ camera), the computed rotation can be used directly to guide it. 
However, here we consider hand-held cameras, where the user is incapable of following such instructions.
This is true regardless of the manner in which the rotation is computed (SFM, panoramic view, or our method). We propose a user interface superimposed on the live preview of the camera, visually indicating the required rotation.


\paragraph{Applications:}
People often like to share images with others.
Indeed, this is one of the main appeals of Facebook, Instagram, etc. Our method allows  a photographer to direct the attention of others present at the same scene to an interesting region, rather than sending an image of that region. This may be more rewarding since the observer can capture his own image of the region, or follow events that take place in that region. It is applicable without any verbal or physical communication, for people who may be at different positions in the scene. The additional images of the scene can be captured by a crowd present at the scene or downloaded from available public photo collections. In many scenarios, however, a public photo collection is not available; hence a short preprocessing time is essential. A similar application is for tourism with a virtual guide. Our method can be used to direct the tourist to rotate his camera to view the section of the scene to which the guide refers. 

Our method can also be used in a homing application that helps people find their friends in a crowd (when GPS or WiFi based localizations are unavailable). Homing algorithms have already been developed for robotics applications, but overlapping fields of view are required (e.g.,~\cite{basri1999visual,goedeme2007omnidirectional}).  Our method can be used to obtain such views. 

A novel setup that we envision is "collaborative photography", where a group of photographers attending the same event cooperate for solving a given task. For example, to compute high quality 3D structure of the scene, many overlapping images from different poses are desired. Our method can be used to obtain additional images on request (see the experiment in Sec.~\ref{sec:experiments}). 

\vspace{0.5cm}

The main contributions of the paper are \begin{inparaenum}[(i)]
\item the introduction of a new challenging task, the camera guidance problem, and its efficient solution;
\item 
the novel SOFA scene representation that allows the camera guidance problem to be solved efficiently while avoiding 3D scene reconstruction;
\item a novel visual user interface for smartphone cameras to guide the user to rotate his camera towards a given scene point.
\end{inparaenum}

\section{Additional Related Work}
\label{sec:RelatedWork}


To allow for geometric and spatial reasoning without direct 3D reconstruction, we use the spatial orderings of the scene points, obtained by rank aggregation. Rank aggregation is the problem of finding a full ranking that agrees with multiple (full or partial) rankings, i.e., a consensus of rankings. It was traditionally studied in the context of social choice and voting theory \cite{young1978consistent}, but was used also for biological sequence alignment \cite{boulesteix2009stability} and web page ranking \cite{dwork2001rank}. Recently, it was used to temporally order a collection of images of a dynamic scene \cite{dekel2014photo,dekel2013space}. 
The common rank aggregation problem is known to be NP hard \cite{dwork2001rank}. In our method, we use the Markov chain approximation for rank aggregation, which was proven to be quite effective if the power iteration method is used \cite{dwork2001rank}.

The camera guidance problem seemingly resides in the field of "active vision" \cite{chen2011active}, where the goal is to change the pose of controlled cameras, e.g., cameras mounted on robots or PTZ cameras, to allow the optimization of some objective. 
For example, when the objective is object tracking, fixation on objects over time is maintained through control of the camera pose. A general approach for the simultaneous tracking of multiple moving targets using a generic active stereo setup is studied by Barreto et al.~\cite{activeTrackingBarreto}. 
As far as we know, none of the methods in the active vision field considered a set of casually taken photographs as in our scenario; moreover, the environment is usually strictly controlled in advance and then manipulated \cite{bodor2007optimal}. Another drawback of these methods is the computational time, where full 3D reconstruction of the scene is usually given in advance, or exhaustively calculated. Our method may be also applied to robot collaboration.

\section{Method}
\label{sec:method}

The input to our method is a pair of images, an initial image $\IC{0}$ captured by camera $\CAM$, and a destination image~$\ID{d}$. In addition, a set of images of the scene, $\setD=\{\ID{j}\}_{\scriptscriptstyle j=1}^{\scriptscriptstyle n}$, possibly captured at (roughly) the same time, is available. The objective is to determine the rotation of $C$ such that a scene point $P_{\scriptscriptstyle T}$ projected to the center region of $I_d$ will also be projected to the center region of a new image captured by $C$, $I^m$. To do so, it is sufficient to compute, $p_{\scriptscriptstyle T}^0$, the projection of $P_{\scriptscriptstyle T}$ to the image plane of $I_0$ (not necessarily in the current FOV of $C$). A visual user interface is then used to assist the photographer in rotating $C$ such that $P_{\scriptscriptstyle T}$ is projected to the center region of $I^m$ (see Sec.~\ref{sec:centralize}).

Before we describe our method, we discuss two alternative methods and their limitations. A direct way to compute $p_{\scriptscriptstyle T}^0$ is by first recovering the projection matrices and the set of scene points, $\setP$, using  structure from motion (SFM) methods (e.g., \cite{snavely2008modeling}). 
This solution is not applicable for online computation due to the long running time required, as we demonstrate in Sec.~\ref{sec:experiments}. Another alternative is to map all images to $I^0$ using homography transformations. In this case, the location of $p_{\scriptscriptstyle T}^0$ may be obtained by $\tilde{p}_T^0 = H\tilde{p}_T^d$ ($\tilde{p}$ is the homogenous coordinates of $p$), where $H$ is the homography between $I^0$ and $I_d$. However, when the scene contains non-negligible 3D structure with respect to the distance from the camera (not planar) or the cameras are separated by considerable translation, a homography transformation does not exist, as discussed in Sec.~\ref{sec:experiments}. 

Instead, we propose an efficient method to compute $p_{\scriptscriptstyle T}^0$  that avoids 3D reconstruction and whose preprocessing time is in the order of minutes rather than hours as in SFM. We first consider the basic case where ${\IC{m}}$ can be captured in  a single step. When the gap between $I^0$ and $I_d$ is large,  intermediate images have to be captured in order to reach ${\IC{m}}$ (see Sec.~\ref{sec:multi}). Then we describe the user interface for guiding the camera rotation.

\subsection{Basic Case: Single Step}
\label{sec:guidance}

We begin with the case that $p_{\scriptscriptstyle T}^0$ is computed in a single step.
We propose to use a {\em supporting subset} of images, ${\cal I}_{\scriptscriptstyle T}^0\subseteq {\cal I}$, that satisfies the following constraints: \begin{inparaenum}[(i)] \item each of the images $I_j\in {\cal I}_{\scriptscriptstyle T}^0$ has sufficient overlap with $I^0$, that is, there are enough corresponding features to compute the fundamental matrix, $F_{0j}$, between them; \item the point $p_{\scriptscriptstyle T}^j$ is detected in each of the images $I_j\in {\cal I}_{\scriptscriptstyle T}^0$. \end{inparaenum} 
Given ${\cal I}_{\scriptscriptstyle T}^0$, the epipolar line that corresponds to $p_{\scriptscriptstyle T}^j$ in image $I^0$ is given by $\tilde{\ell}_j = F_{0j}\tilde{p}_{\scriptscriptstyle T}^j$. The epipolar point transfer (EPT) \cite{hartley2003multiple} is used to compute $p_{\scriptscriptstyle T}^0$, that is, the intersection of a pair of epipolar lines, $\ell_j$ and $\ell_k$. Note that $p_{\scriptscriptstyle T}^j$ is not necessarily within the FOV of $I_0$. For robustness, when $|{\cal I}_T^0|>3$, the intersection point with the most epipolar line inliers is found, and the outliers are discarded. When the number of epipolar lines is large, the RANSAC algorithm \cite{fischler1981} is used. In our method we typically restrict $|{\cal I}_T^0|<5$.

\subsection{General Case: Multiple Steps}
\label{sec:multi}

When the gap between $I^0$ and $I_d$ is large, the supporting set of images, ${\cal I}_T^0\subseteq {\cal I}$, does not exist and the EPT cannot be used directly to compute $p_{\scriptscriptstyle T}^0$. Hence, we suggest rotating $C$ to a sequence of intermediate views, $I^1, \ldots, I^m$, until reaching the desired overlap with~$I_d$. A sequence of points $p_1, \ldots ,p_m$ are chosen such that $C$ is rotated to center $p_k$ in image $I^k$; that is, $p_k$ is at the center region of $I_k$. If $p_m=p_{\scriptscriptstyle T}$ then $I^m$ is the desired final image.


To this end, we define spatial orders between scene points. We next describe how such orders can be used to compute the sequence, $p_1, \ldots,p_m$, and then show how the orders can be efficiently computed (the SOFA representation).

\subsubsection{Sequence Properties}
Let us first assume that the order of the scene points is given by 
the order of the $x$ coordinate of their projections to the image plane of $I^0$ (not necessarily within the FOV of $I^0$). Note that since this is a non-trivial assumption, we show in Sec.~\ref{sec:spatialorder} how this order can be (approximately) computed. Let $p_c^0$ be a point at the center region of $I_0$. The relative rank of $p_{\scriptscriptstyle T}^0$ and $p_c^0$ in the ordered sequence  determines whether to rotate $C$ "to the left" or "to the right".
We next show how this is used to determine the first intermediate image, $I^1$. This is an iterative process that is repeated until the final image, $I^m$, is reached.

Formally, let $S_x$ be the sequence of features ordered by their $x$ coordinate. The spatial order is defined by the feature permutation $\sigma_x$. That is, $\sigma_x(p_k^0)$ is the rank of the feature point $p_k^0$ in $S_x$. The feature ranked $i$-th in $S_x$ is given by $S_x(i)$.
Let $S_x(\gamma)=p_{\scriptscriptstyle T}^0$ and $S_x(\alpha)=p_c^0$. We use the order of points in $S_x$ to choose a new point, $S_x(\beta)$, to be centered in the next image. Assume without loss of generality that $\alpha<\gamma$, that is, $S_x(\alpha)$ precedes $S_x(\gamma)$ in $S_x$. We choose the new point $S_x(\beta)$ 
 such that \begin{inparaenum}[(i)] \item $\alpha<\beta\leq\gamma$ and
$\gamma-\beta$ is minimal; \item a supporting set of images for computing $S_x(\beta)$ exists (see Sec.~\ref{sec:guidance}). \end{inparaenum} Then $C$ is rotated to capture a new image $I^1$ so that $S_x(\beta)$ is at its center region (see Sec.~\ref{sec:centralize}). This procedure is repeated until $S_x(\beta)=p_{\scriptscriptstyle T}^m$ is centered in $I^m$.

In a similar way, let $S_y$ be the sequence of features ordered by their $y$ coordinate and defined by the feature permutation, $\sigma_y$. An additional constraint is added by $\sigma_y$ for choosing a new point, $p$, for centering in $I^1$. 
The feature ranked $i$-th in $S_y$ is given by $S_y(i)$.
Each point $p$ appears in both sequences, $S_x$ and $S_y$, but may have a different ranking. Let $\gamma'$ be the ranking of $p_{\scriptscriptstyle T}^0$ (i.e., $S_y(\gamma')=p_{\scriptscriptstyle T}^0$), and $\alpha'$ be the ranking of $p_c^0$ (i.e., $S_y(\alpha')=p_c^0$).  
Assume without loss of generality that $\alpha'<\gamma'$. We choose a new point, $p=S_y(\beta')=S_x(\beta)$, such that \begin{inparaenum}[(i)] \item $\alpha<\beta\leq\gamma$~,~ $\alpha'<\beta'\leq\gamma'$ and $(\gamma-\beta)+(\gamma'-\beta')$ is minimal; \item a supporting set of images for computing $p$ exists. \end{inparaenum}

\subsection{SOFA Computation}
\label{sec:spatialorder}
The 3D location of the scene points can be used to compute their projections to $I^0$, and hence their orders, $\sigma_x$ and $\sigma_y$. Since we are trying to avoid 3D reconstruction, each 3D point is represented by a set of matched features in the set of images, ${\cal I}$ (see Sec.~\ref{sec:dictionary}). Consider the ideal case where \begin{inparaenum}[(i)] \item a perfect matching between image features is available in all images; \item the order of corresponding points is preserved in all images; and \item there exists sufficient overlap between images. \end{inparaenum} Note that if these three conditions hold, the spatial orders are identical in all images, and define the spatial orders of the corresponding 3D scene points. In this case $\sigma_x$ and $\sigma_y$ are obtained by simply combining the partial orders from all images.

However, in practice there are matching errors and the order of corresponding points is not preserved in all images. To overcome this problem, we define the spatial orders on the 3D scene points, $\hat{\sigma}_x$ and $\hat{\sigma}_y$, that are as consistent as possible with respect to the spatial orderings of their visible projections to the set of images.
Ranking the order of the scene points from a noisy set of rankings is cast as the well-known rank aggregation problem. In our case, each image provides a ranking of the 3D scene points according to the spatial locations of their projections to the image.
We next describe an efficient approximate solution for computing the correspondence between features in a large set of images and an approximate solution to the rank aggregation problem.

\subsubsection{Feature Correspondence via a Visual Dictionary}
\label{sec:dictionary}
An approximate feature correspondence can be obtained by using a dictionary of visual words (e.g., \cite{sivic2004video}), where each bin, $B_i$, represents the projection of the same scene point, $P_i\in{\cal P}$. In the rest of this section, we regard $B_i$ as the representative of the projection of a scene point $P_i$.
A straightforward (but costly) alternative is to compute a pairwise matching between each pair of images (e.g., \cite{lowe2004distinctive}). 
Using a dictionary allows us to overcome the time complexity involved in pairwise matching (at least $O(n^2))$. Its robustness is sufficient for our method, as demonstrated experimentally (see Sec.~\ref{sec:experiments}). In our implementation we use SIFT features \cite{lowe2004distinctive} that are clustered using hierarchical K-means \cite{vedaldi08vlfeat}.

\subsubsection{Rank Aggregation}
\label{sec:rank}

The widely accepted objective to minimize in rank aggregation is the \emph{Kendall distance}, that is, the pairwise disagreements between the full order and each of the partial orders computed in each image. Formally, the Kendall distance, $K(\sigma,\sigma_i)$, between the full order, $\sigma$, and a partial order, $\sigma_i$, of the sequence, $S_i$, is defined by
\[
 K(\sigma,\sigma_i) =  
\sum_{\substack{l,k\in S_i \\ \sigma_i(l)<\sigma_i(k)}}{b_>(\sigma(l),\sigma(k))},
\]
where $b_>(i,j)=1$ if $i > j$ and $b_>(i,j)=0$ otherwise.

The rank aggregation problem is formally defined by minimizing
\[
\sigma^* = \argmin\limits_\sigma\sum_i^{|\mathcal{I}|}{K(\sigma,\sigma_i)}.
\]
Since minimizing this objective was proven to be NP-hard, we use the Markov chain approximation to this problem~\cite{dekel2014photo,dwork2001rank}. In our method we employ two rank aggregation instances for the $x$ and $y$ coordinates independently. We next briefly describe the Markov chain approximation for rank aggregation for the $x$ coordinate.

Let $G = (V,E,w)$ be a weighted and directed graph. A node $v_i\in V$ corresponds to the bin $B_i$. The weight, $w(e)$, of a directed edge, $e=(v_i,v_j)\in E$, corresponds to the vote that $\hat{\sigma}_x(B_i)<\hat{\sigma}_x(B_j)$. It is computed based on the spatial distances of the image points $p_i^k\in B_i$ and $p_j^k\in B_j$ in the image $I_k\in \mathcal{I}$. That is,
\[
w(e)=\sum\limits_{I_k\in \mathcal{I}}{x(p_j^k)-x(p_i^k)},
\]
where $x(p)$ is the $x$ coordinate of $p$ and $x(p_i^k) < x(p_j^k)$. To resolve conflicts in the order between $B_i$ and $B_j$, e.g., due to correspondence errors, the directed edge, $e_{ij}=(v_i,v_j)$ or $e_{ji}=(v_j,v_i)$, with the smaller weight is discarded; that is, if $w(e_{ij})>w(e_{ji})$ then we keep $e_{ij}$.


\begin{figure*}[t]
		\begin{tabular}{c c c c}
			\includegraphics[width=0.245\linewidth]{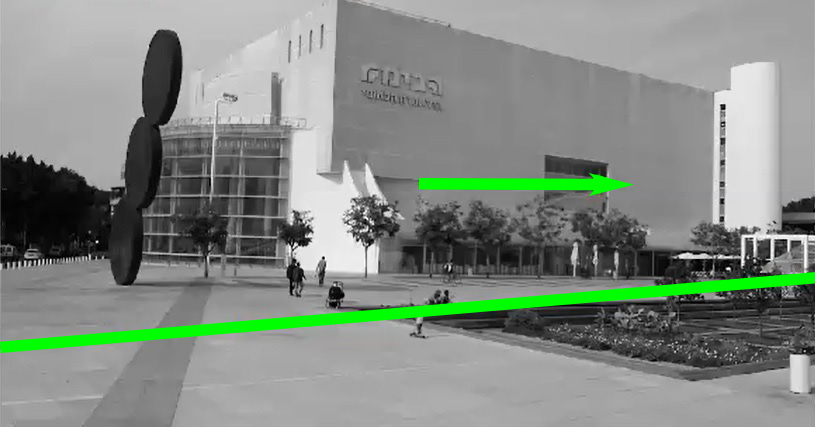} & \includegraphics[width=0.245\linewidth]{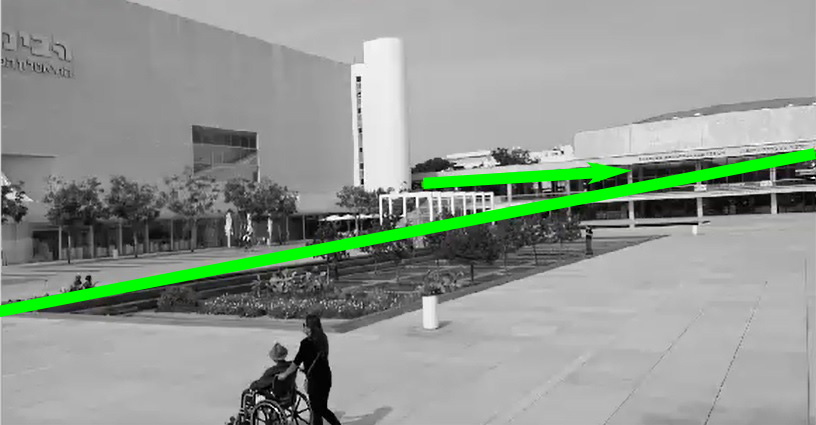} & \includegraphics[width=0.245\linewidth]{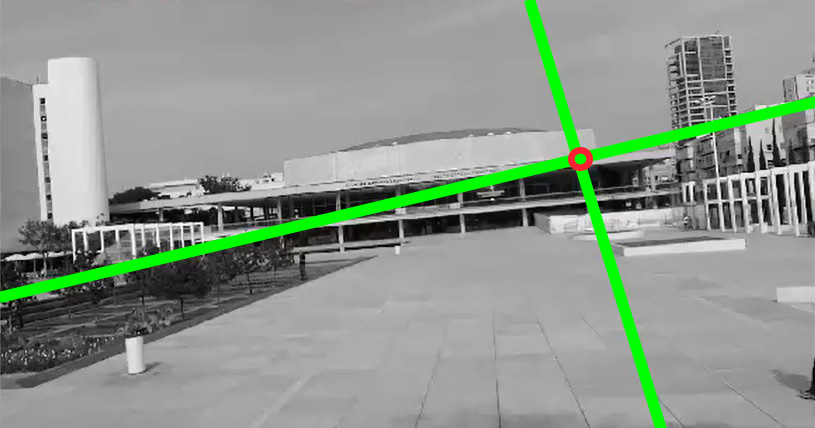} & \includegraphics[width=0.245\linewidth]{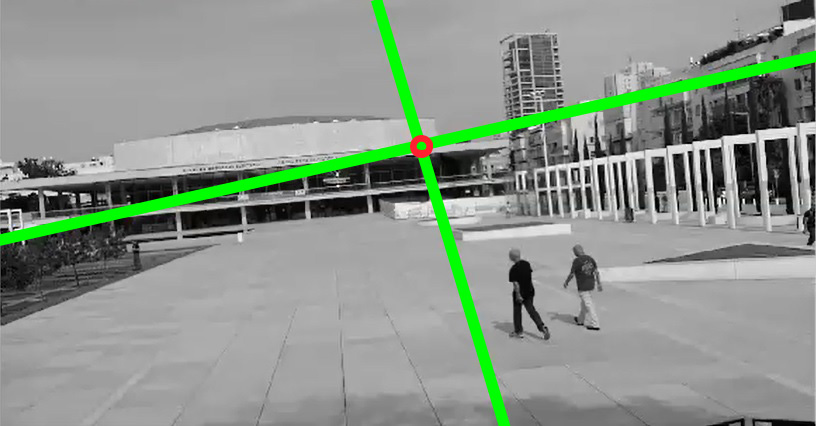} \\
			(a) & (b) & (c) & (d)
			\end{tabular}
		\caption{The user interface for rotating the camera, $C$, to center a point, $p_{\scriptscriptstyle T}^i$, in the intermediate view, $I^i$. In (a) and (b), $p_{\scriptscriptstyle T}^i$ is out of the FOV so an arrow indicates its direction. In (c) and (d), $p_{\scriptscriptstyle T}^i$ is within the FOV and marked by a red circle. In (d), $p_{\scriptscriptstyle T}^i$ is in the center region of $I^{i+1}$. The images were taken from dataset \emph{urban2}.}
\label{fig:figure3}
\end{figure*}

\begin{figure}[t]
  \centering
		\def\svgwidth{230pt}
    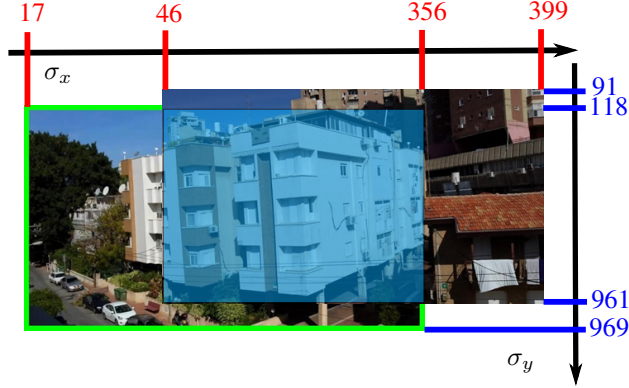
  \caption{Intervals defined by the SOFA representation. The intervals of the left image are [17,356] and [118,969] in $\sigma_{x}$ and $\sigma_{y}$, respectively. The intervals of the right image are [46,399] and [91,961] in $\sigma_{x}$ and $\sigma_{y}$, respectively. The overlap between the two images is defined as the overlap of their intervals. That is, [46,356] and [118,961] in $\sigma_{x}$ and $\sigma_{y}$, respectively.}
\label{fig:figure11}
\end{figure}

The Markov chain approximation is defined by a graph of states and a transition matrix, $M$, that corresponds to the probability of transition from one state to the other. The idea is that a sufficient number of steps in a random walk over the graph will end up in the state that corresponds to the element to be ranked last. If this state is removed, the process may be repeated until all elements are ranked. In our method, the set of states is defined to be $V$ and $M_{ij}=w(e)/n_e$, where $e=(v_i,v_j)$ and $n_e$ is a normalization constant so that the sum of each row in $M$ is exactly $1$. Assuming a uniform probability distribution over the $|V|$ states, defined by a vector $x$, the probability distribution after a random walk of $k$ steps is $M^{k}x$. The random walk eventually converges to the eigenvector $y = My$. In our method we obtain a good approximation of $y$ by running a few power iterations until a steady state is reached. The state with the highest probability from $y$ is removed and the process is repeated until the full spatial ordering, $\sigma_x$, is obtained.

The expected limitation of using the dictionary for computing feature correspondence is false positive matching, which may affect the rank aggregation results. To reduce the number of false positives, a large dictionary is used (see~Sec.~\ref{sec:details}). The other source of errors is that the order of the scene points is not preserved in the set of images, since it consists of objects with different depths, for example a tree or a pole. In practice the average Kendall distance between the global rank and the local rank in each image on the sets we considered is small ($\sim$5\%).
 Moreover, for the application at hand, our method is able to deal with these errors successfully, as we show in Sec.~\ref{sec:experiments}.

\subsubsection{Sequence Construction}
\label{sec:seqConstruction}
The SOFA representation is used for determining the rotation direction and in particular the supporting set of images. We compute an interval in the global ranking for each image. The overlap between two images is defined by the overlap between two such intervals (see Fig.~\ref{fig:figure11}). To overcome ranking errors, the interval of each image is computed using the medians of the first and the last deciles.


We will now show how to choose a scene point $P$, which is represented by a bin $B$,
so that its projection is centered in the next intermediate image, $I^{i+1}$. Let $S_x$ and $S_y$ be the $x$ and $y$ rankings of the dictionary bins $\{B_i\}$, computed using rank aggregation as described above. 
The bins $B_{\scriptscriptstyle T}$ and $B_c$ correspond to projections of the scene points, $P_{\scriptscriptstyle T}$ and $P_c$, that are viewed in the center regions of $I_d$ and $I^i$, respectively. The bin $B_{\scriptscriptstyle T}$, and similarly $B_c$, has a different rank in $S_x$ and in $S_y$. Let $\hat{\sigma}_x(B_{\scriptscriptstyle T})=\gamma$, $\hat{\sigma}_y(B_{\scriptscriptstyle T})=\gamma'$, $\hat{\sigma}_x(B_c)=\alpha$ and $\hat{\sigma}_y(B_c)=\alpha'$. We define for each $B$ for which $\sigma_x(B)\in[\alpha,\gamma]$ and $\sigma_y(B)\in[\alpha',\gamma']$ the value $d(B)=|\hat{\sigma}_x(B)-\gamma|+|\hat{\sigma}_y(B)-\gamma'|$. A list ${\cal L}$ is then obtained by sorting $d$ in increasing order. We then traverse ${\cal L}$ to find a $B$ for which 
a sufficiently large supporting set of images exists, in which features $p\in B$ are detected. Once such $B$ is found, we apply the basic case solution (Sec.~\ref{sec:guidance}), where the fundamental matrices of the support set of images with respect to $I^i$ are computed using the BEEM algorithm \cite{goshen2008balanced}. It receives as input the initial correspondences using the dictionary (Sec.~\ref{sec:dictionary}) and is able to overcome correspondence errors.

\begin{figure*}[t]
\begin{tabular}{c c c c c}
						\includegraphics[width=0.195\linewidth]{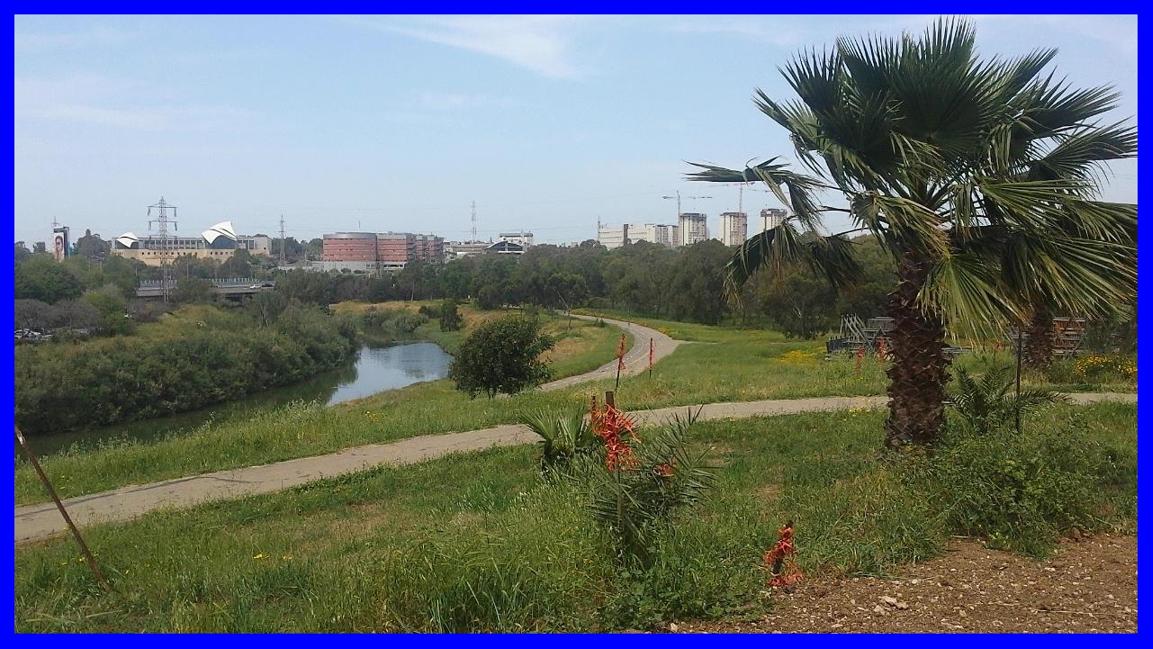} &\includegraphics[width=0.195\linewidth]{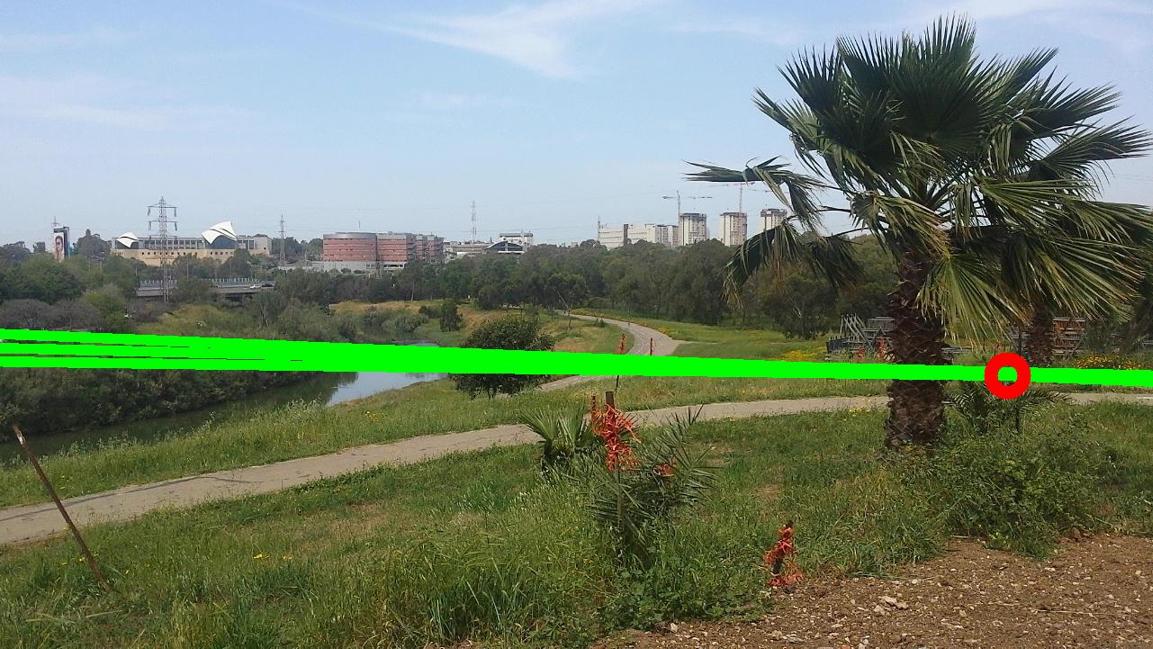}& \includegraphics[width=0.195\linewidth]{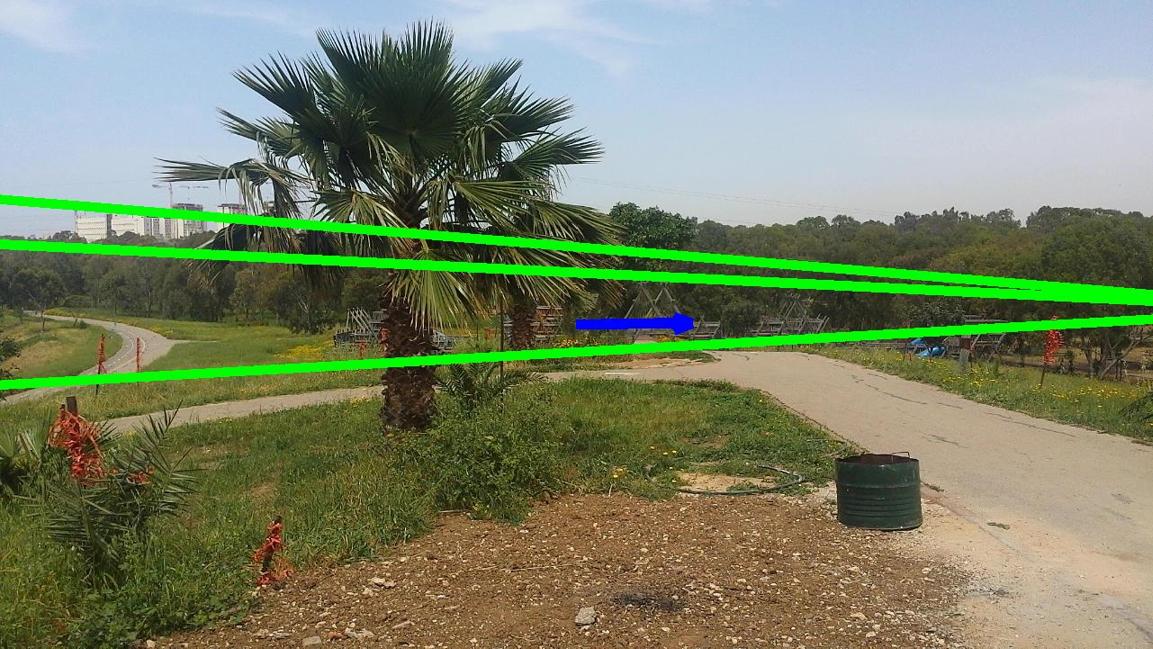}& \includegraphics[width=0.195\linewidth]{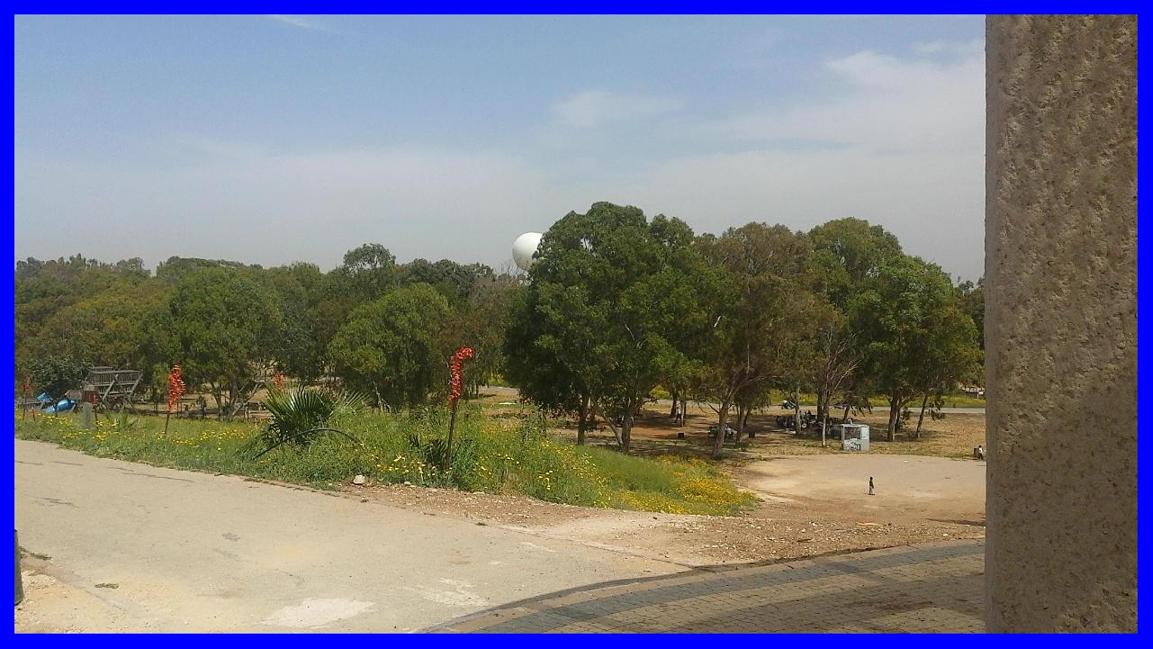}& \includegraphics[width=0.195\linewidth]{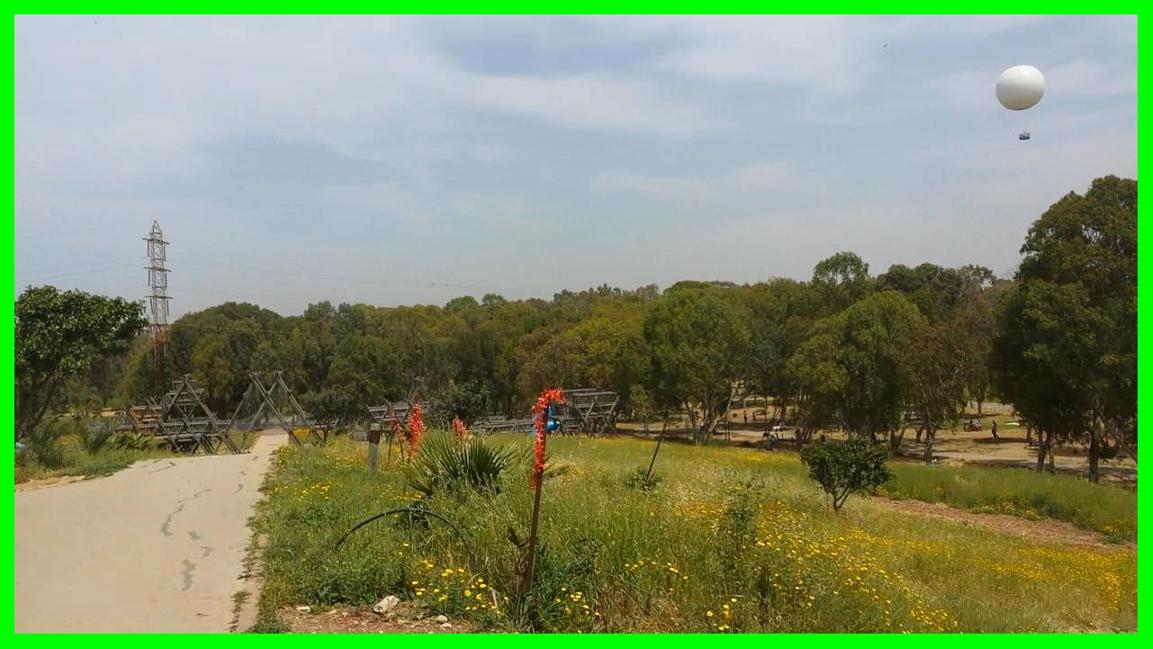} \\
			\includegraphics[width=0.195\linewidth]{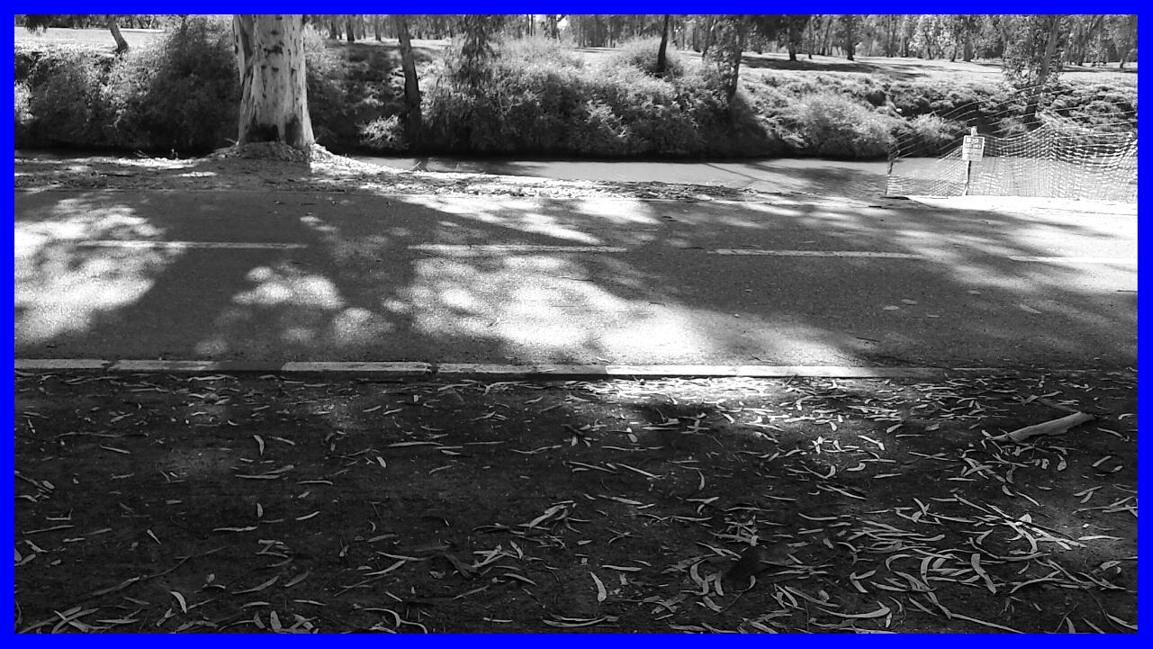} &\includegraphics[width=0.195\linewidth]{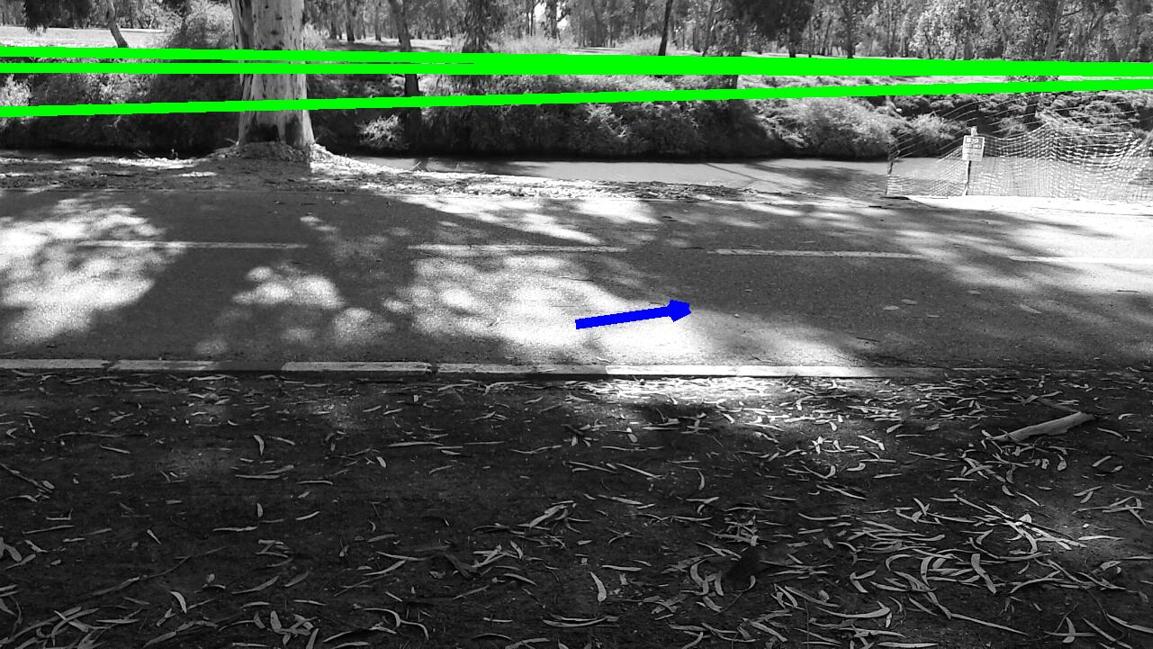}& \includegraphics[width=0.195\linewidth]{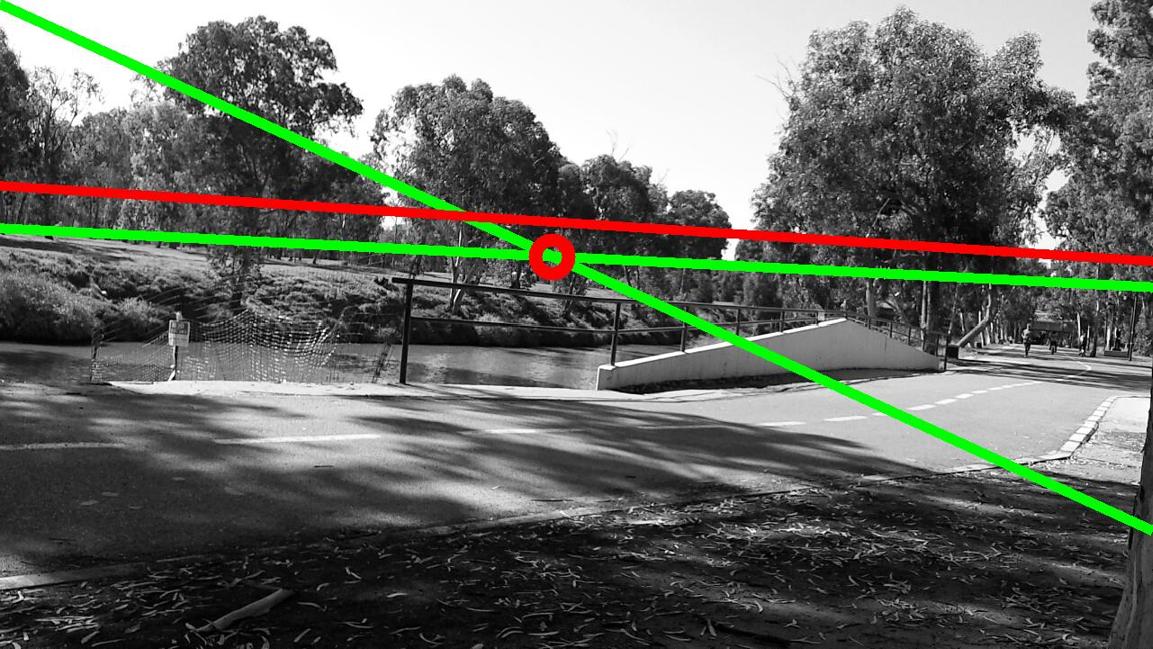} &\includegraphics[width=0.195\linewidth]{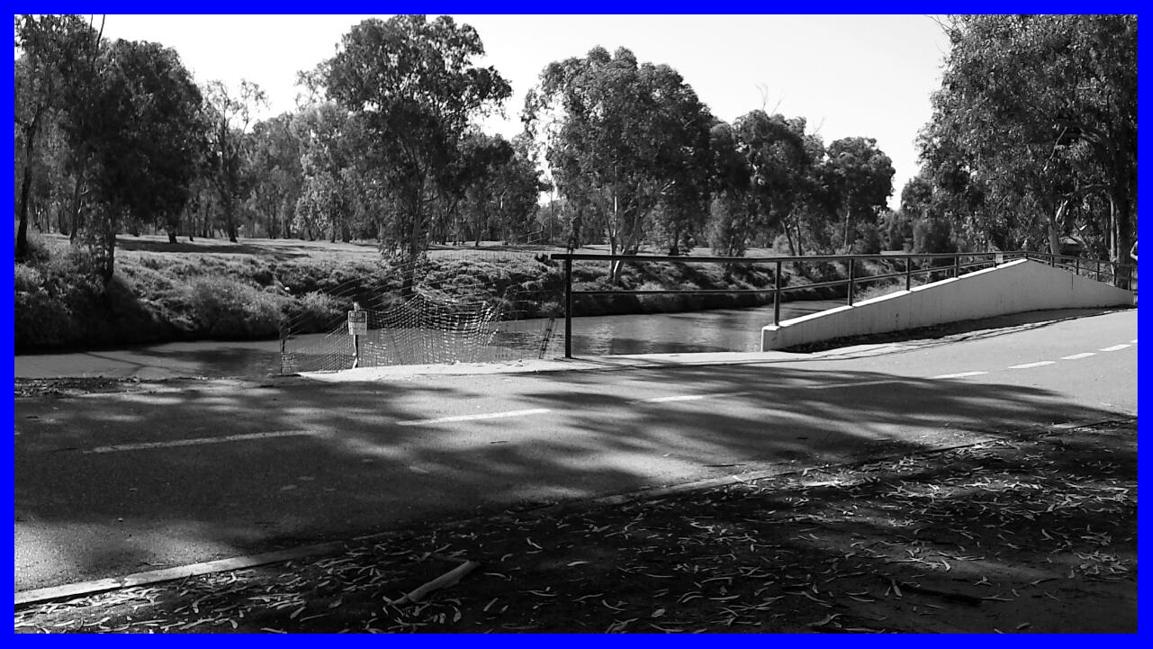}& \includegraphics[width=0.195\linewidth]{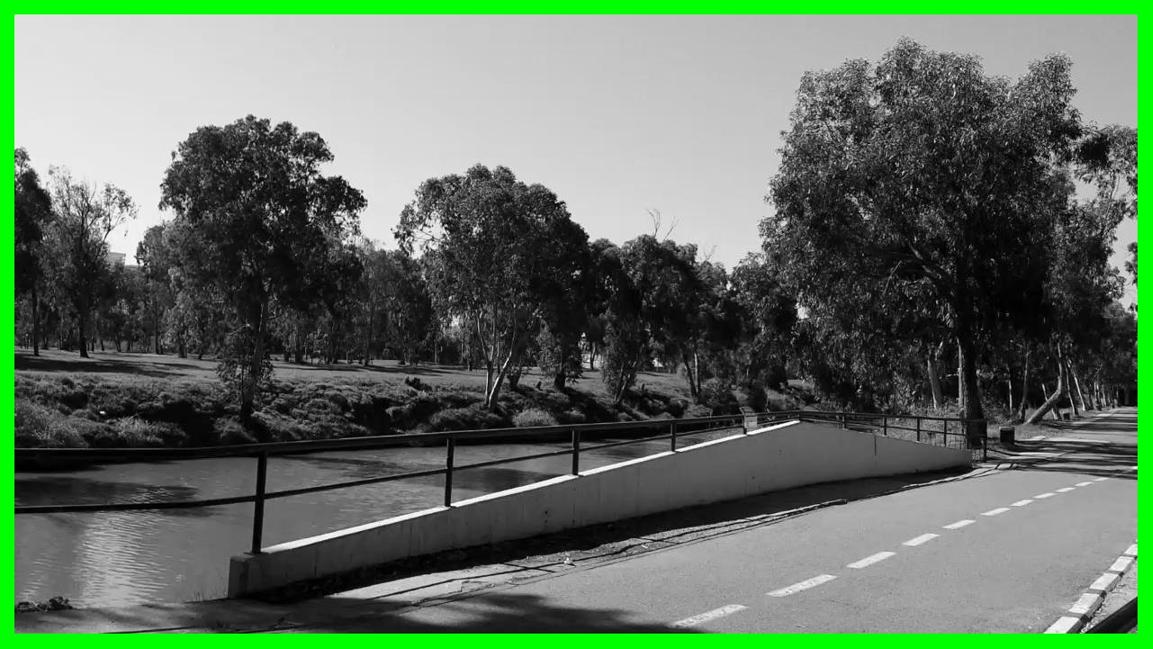} \\
			\includegraphics[width=0.195\linewidth]{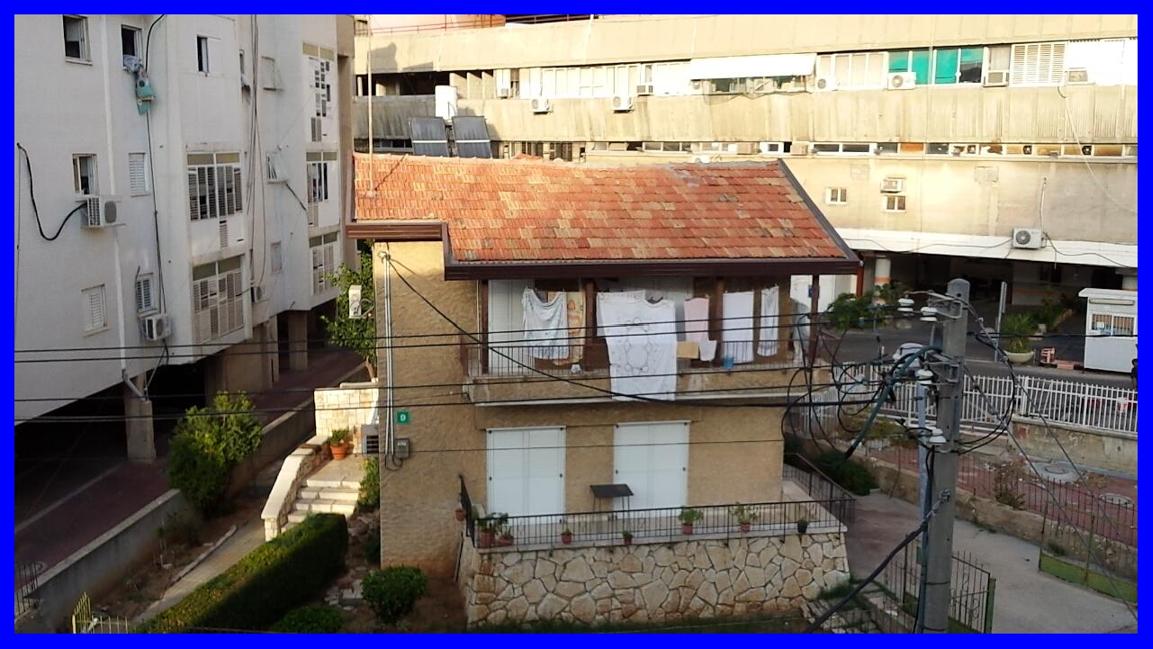} &\includegraphics[width=0.195\linewidth]{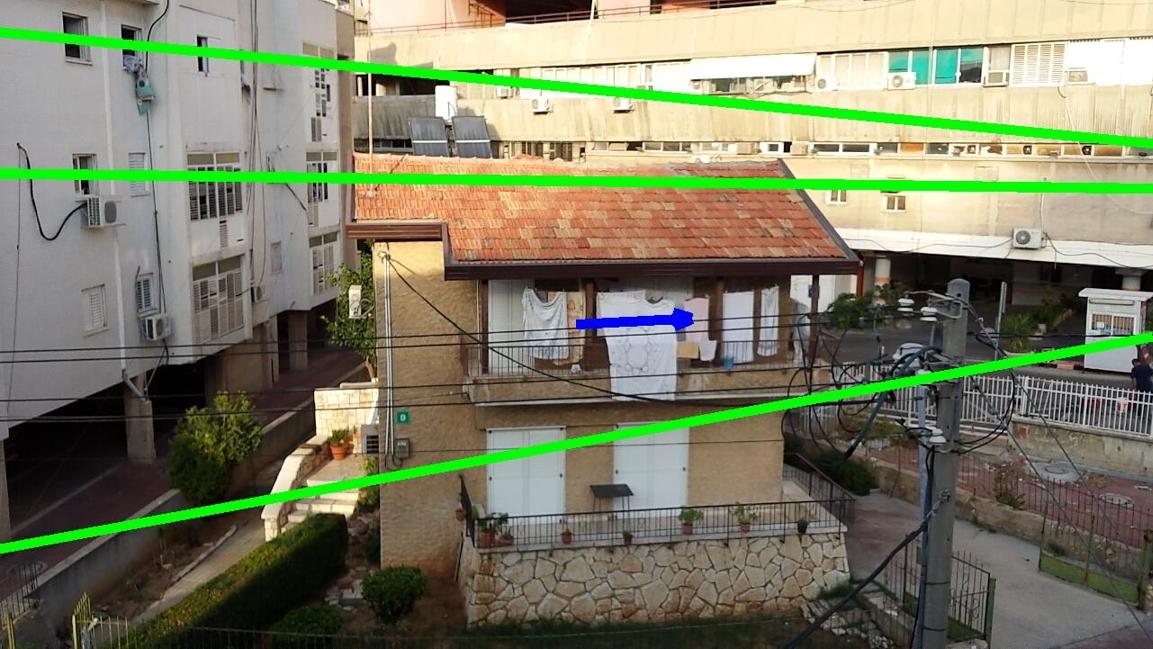}& \includegraphics[width=0.195\linewidth]{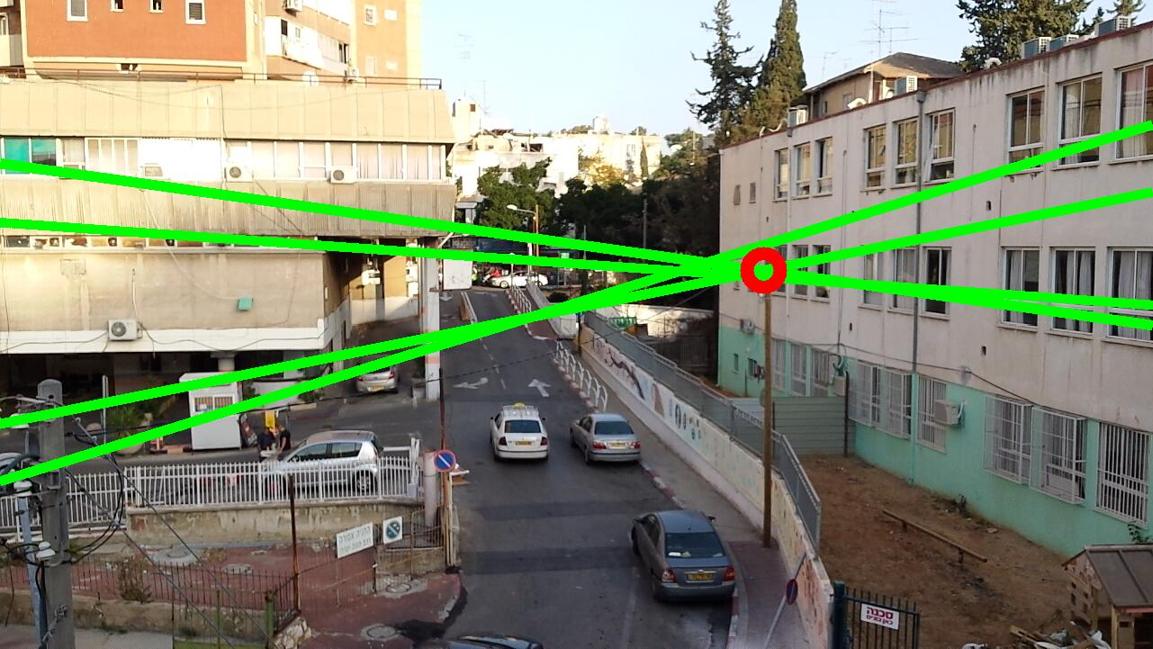} &\includegraphics[width=0.195\linewidth]{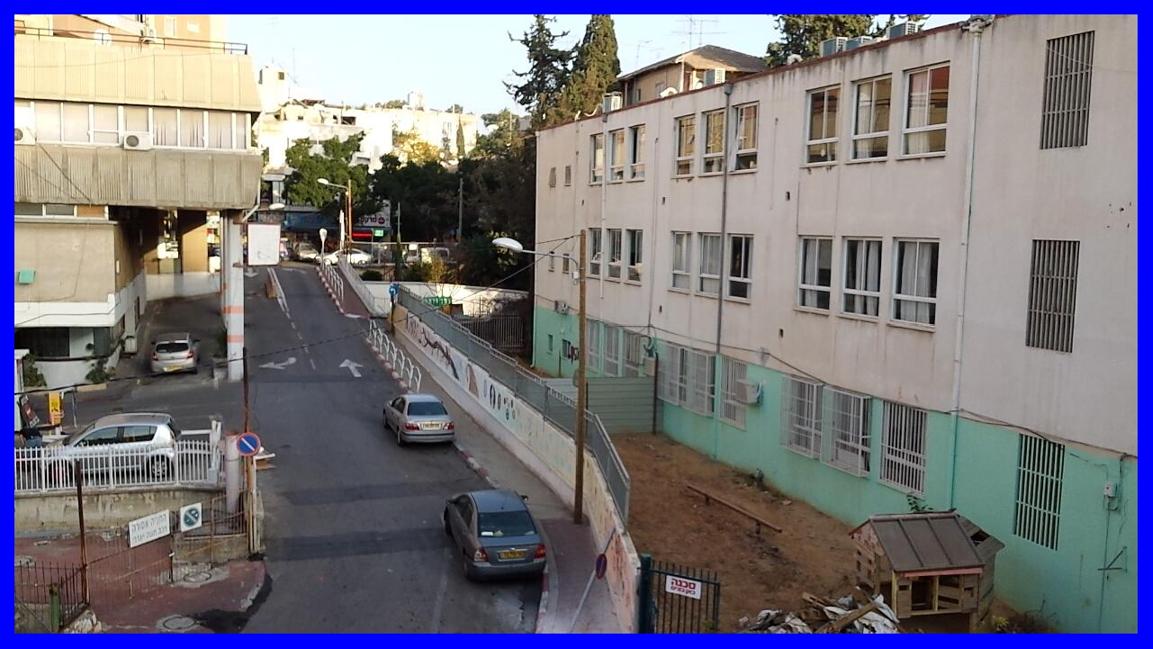}& \includegraphics[width=0.195\linewidth]{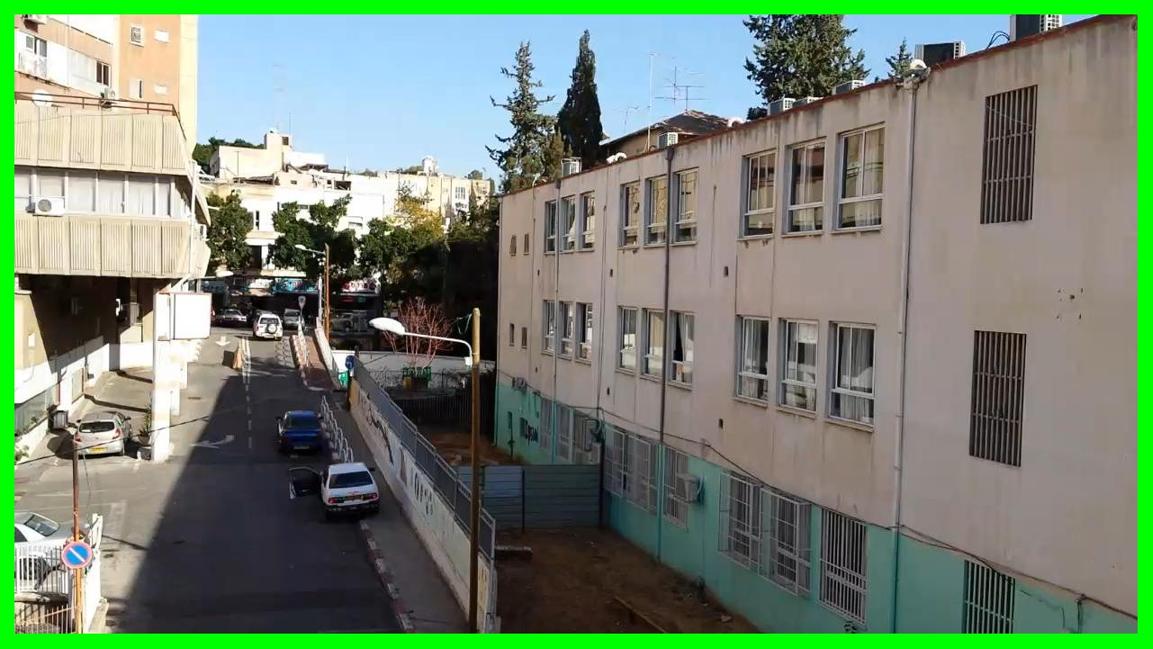} \\
		$I^0$ &  $I^0$ + interface  & $I^1$ + interface &  $I^m$   & $I_d$ \\	
\end{tabular}
\caption{Camera guidance examples. In the top row -- dataset \emph{park1}; in the middle row -- dataset \emph{park2}; in the bottom row -- dataset \emph{urban1}. Note that the green lines correspond to inlier epipolar lines, while the red lines correspond to outliers (see Sec.~\ref{sec:guidance}).}
\label{fig:figure1}
\vspace{-0.2cm}
\end{figure*}

\subsection{Visual User Interface}
\label{sec:centralize}

In this section we describe the visual user interface that assists a photographer in the required rotation of camera $C$. Let $p$ be the projection of a scene point $P$ to the image plane of $I$ captured by $C$.
We propose a user interface that is superimposed on top of the live preview of the camera. We wish to notify the photographer of the location of $p$ in the image plane of every frame $f$ in the live preview. 
When $p$ is within the FOV of $f$, then it is marked on the frame (Fig.~\ref{fig:figure3}(c)). Once $p$ is marked, it is easy for the photographer to rotate the camera to center it. When $p$ is outside the FOV of $f$, only the direction from the center of $f$ to $p$ is marked (e.g., using an arrow (Fig.~\ref{fig:figure3})). To further assist the photographer, the epipolar lines are marked on $f$ so the photographer knows that $p$ is at the intersection of their extensions.

While the user rotates $C$, the marked location of $p$ (or the directing arrow or the set of epipolar lines) is updated. For the first frame, the location of $p$ as well as the pair of epipolar lines are computed using the supporting set of images (see Sec.~\ref{sec:guidance}). Since the user is asked to rotate the camera, we can assume that two frames of the live preview, $f_i$ and $f_j$, are related by a homography transformation, $H_{i,j}$. Homography transformations can be composed; hence, the homography, $H_{0,j}$, between any frame, $f_j$, and $f_0=I$, can be computed. The updated location of $p$ in frame $f_j$ is given by $\tilde{p}_j=H_{0,j}\tilde{p}$, and the updated equation of an epipolar line $\tilde{\ell}$ is given by $\tilde{\ell}_j=H_{0,j}^{-T}\tilde{\ell}$.
The homography $H_{i,j}$ can be computed using RANSAC on a set of corresponding points in the two frames. For implementation details see Sec.~\ref{sec:details}.

Our method can also be used when the camera is automatically controlled (e.g., a robot or a PTZ camera). In this case the interface is much simpler since the rotation can be directly conveyed using an axis and angle. Given the internal calibration matrix, $K$, that corresponds to image $I$, the rotation axis, $\hat{a}$ and angle, $\theta$, for centering $p$ are given by \cite{hartley2003multiple}
\[
\begin{array}{lcl}
\theta = \arccos(\hat{d}(\textbf{0})^T\cdot\hat{d}(p)), \\
\hat{a} = \hat{d}(\textbf{0})\times\hat{d}(p),
\end{array}
\]
where $\hat{d}(p) = K^{-1}\tilde{p}^T/||K^{-1}\tilde{p}^T||$ is the normalized direction that corresponds to $p$, and \textbf{0} is the center of the image.

\begin{figure*}[t]
		\begin{tabular}{c c c c c}
			\includegraphics[width=0.195\linewidth]{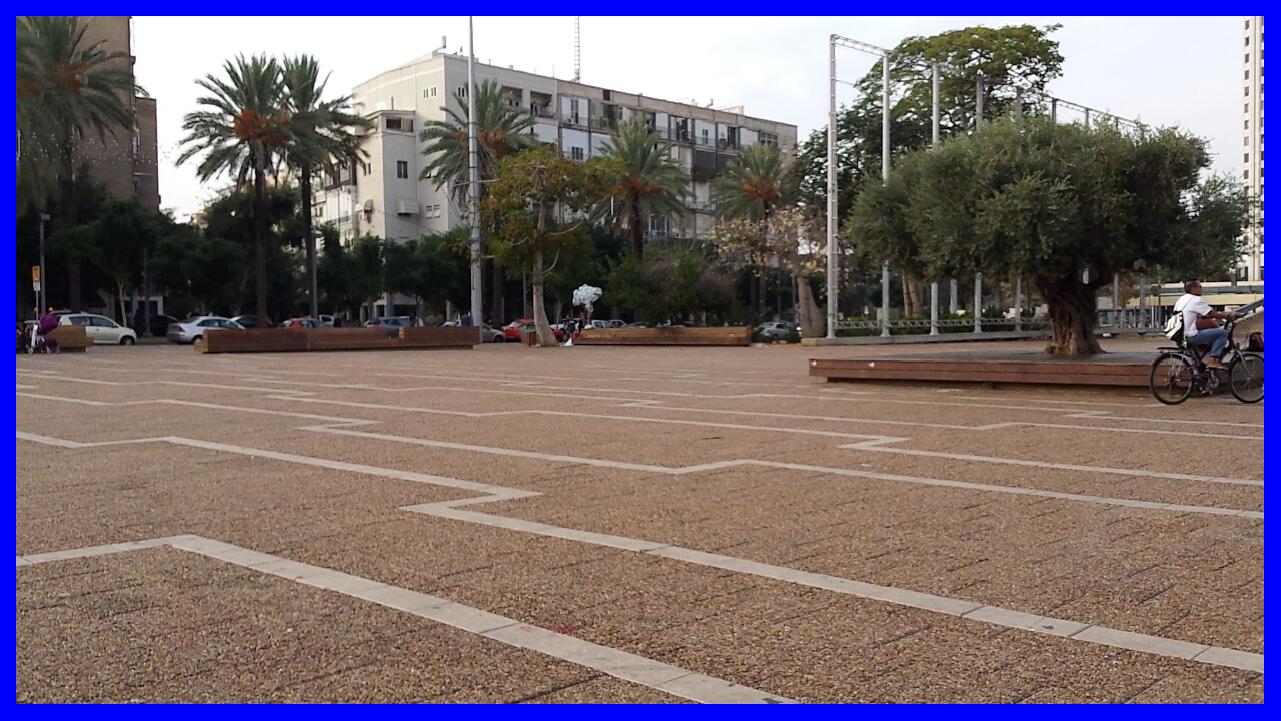} & \includegraphics[width=0.195\linewidth]{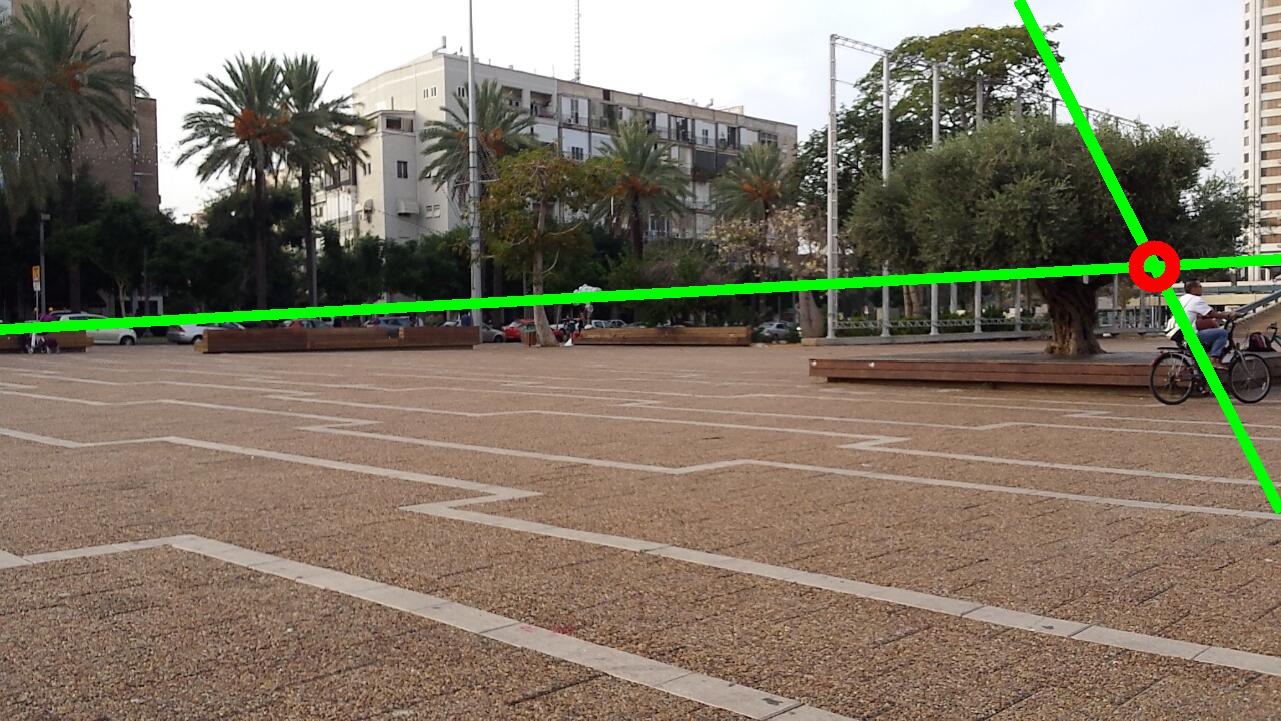} & \includegraphics[width=0.195\linewidth]{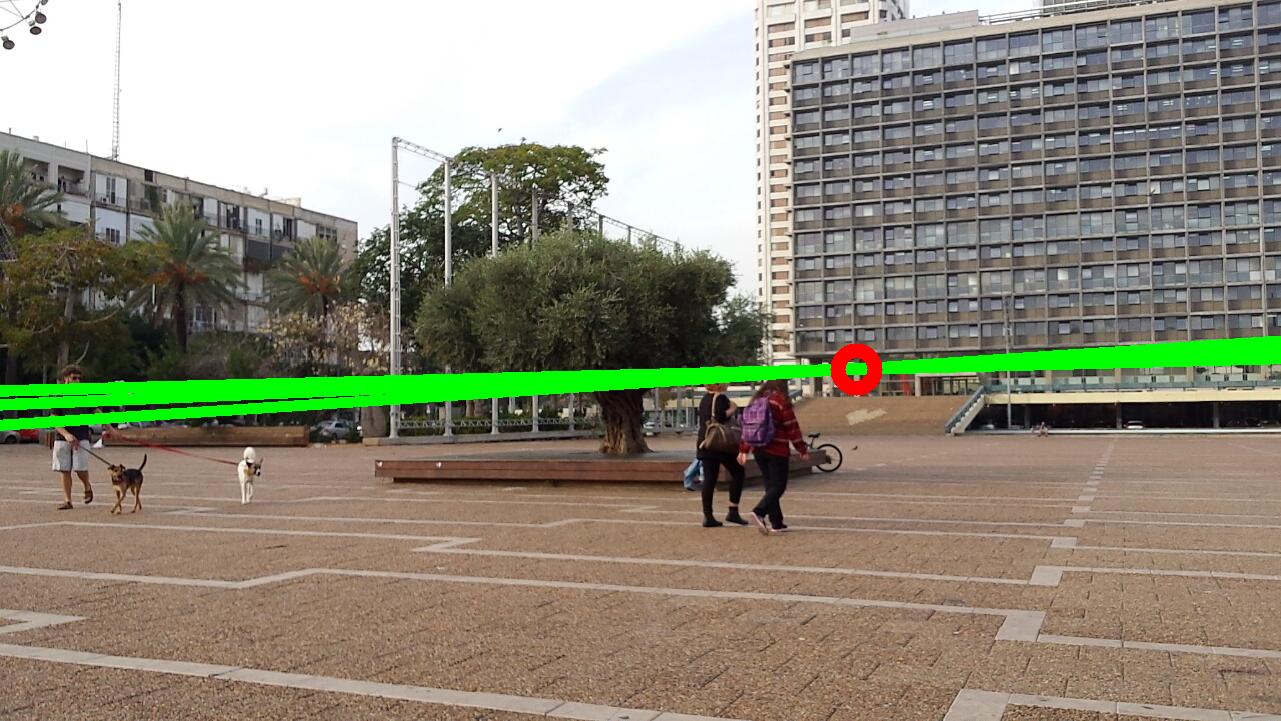} &\includegraphics[width=0.195\linewidth]{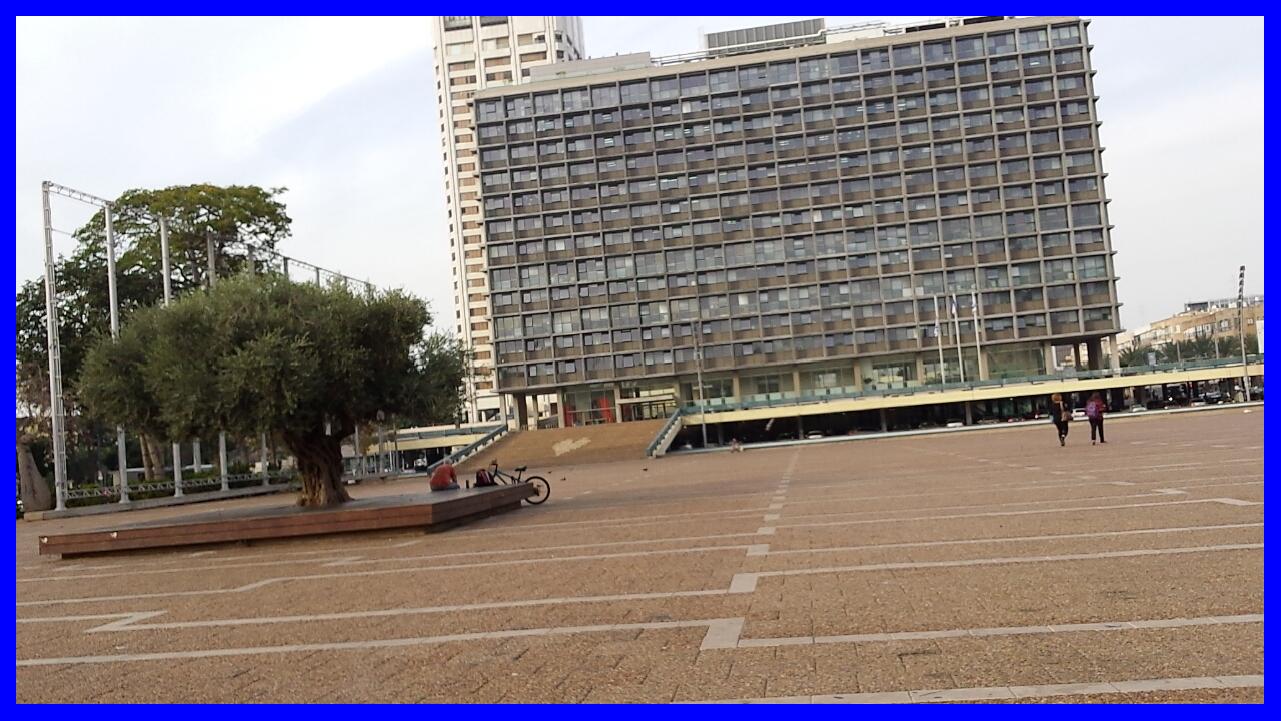} & \includegraphics[width=0.195\linewidth]{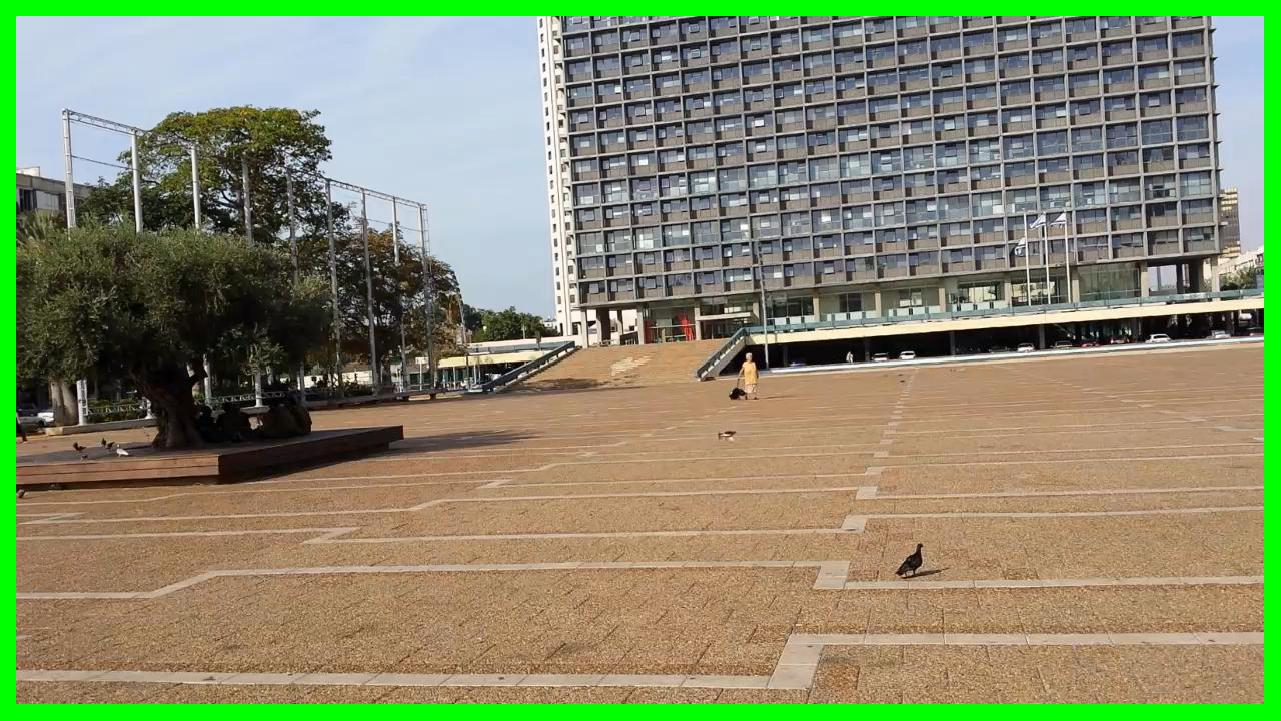} \\

			$I^0$&  $I^0$ + interface  & $I^1$ + interface &  $I^m$   & $I_d$ \\
			\includegraphics[width=0.195\linewidth]{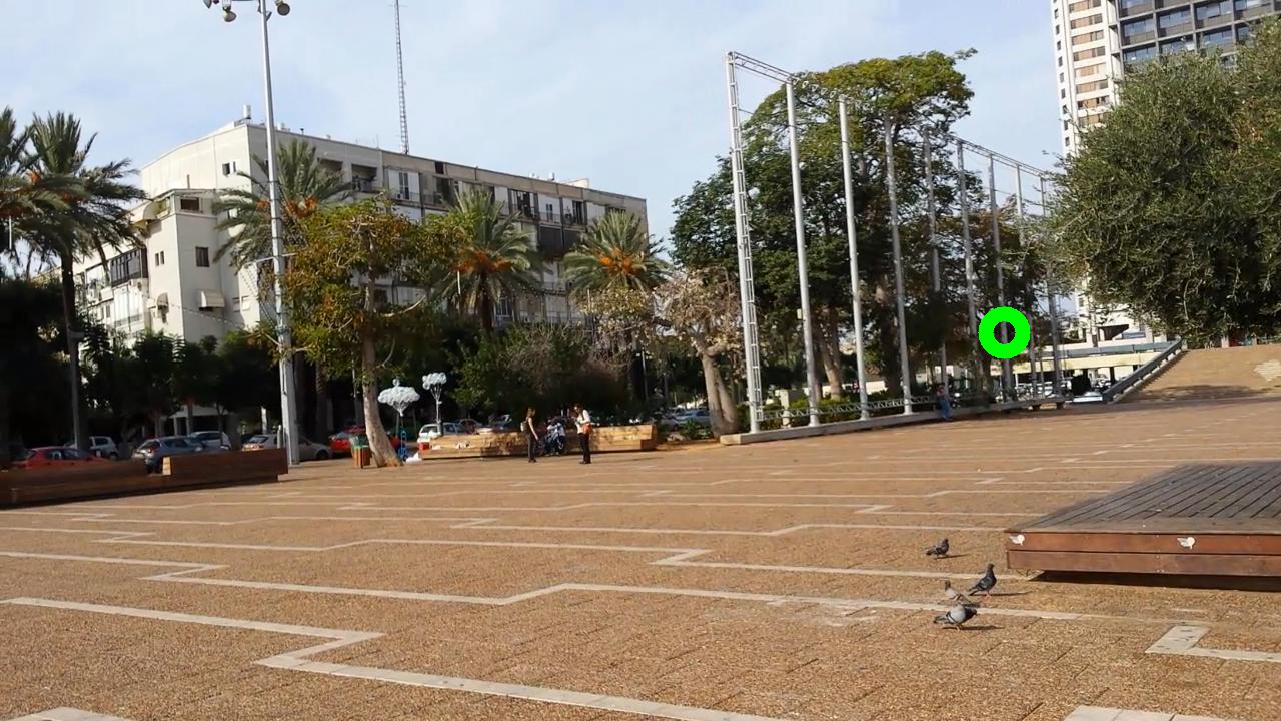} &\includegraphics[width=0.195\linewidth]{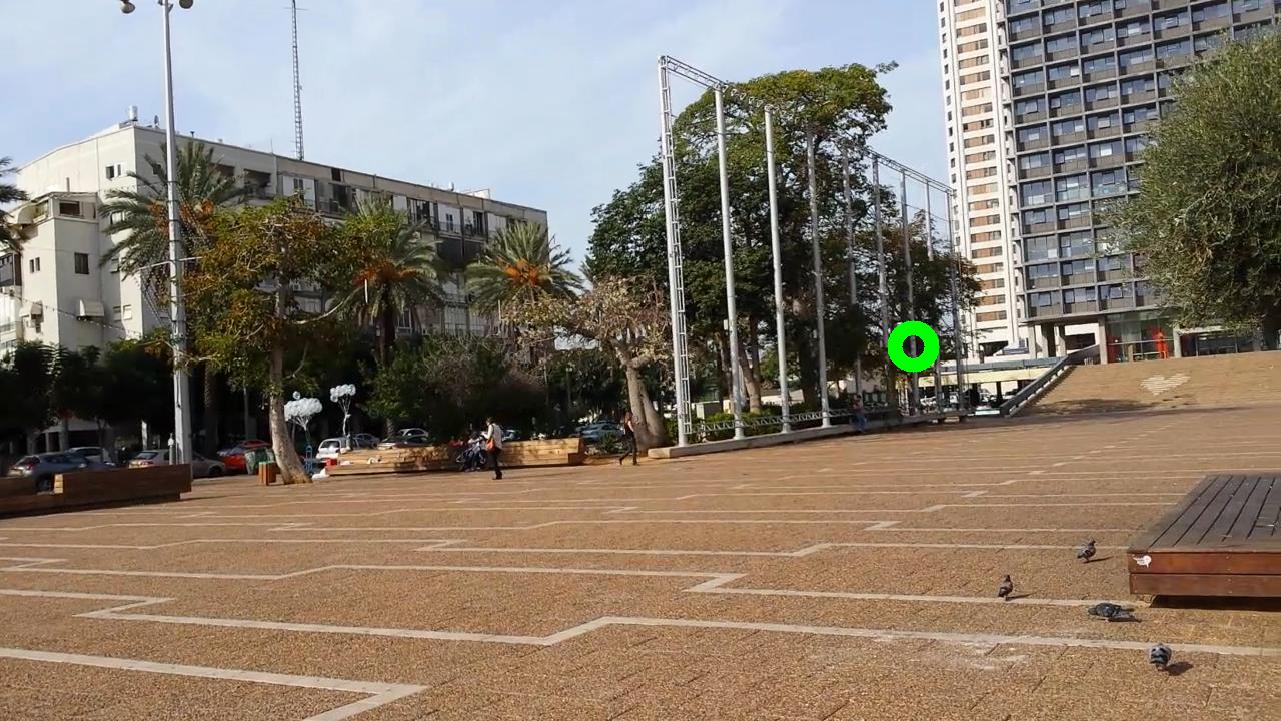}&	\includegraphics[width=0.195\linewidth]{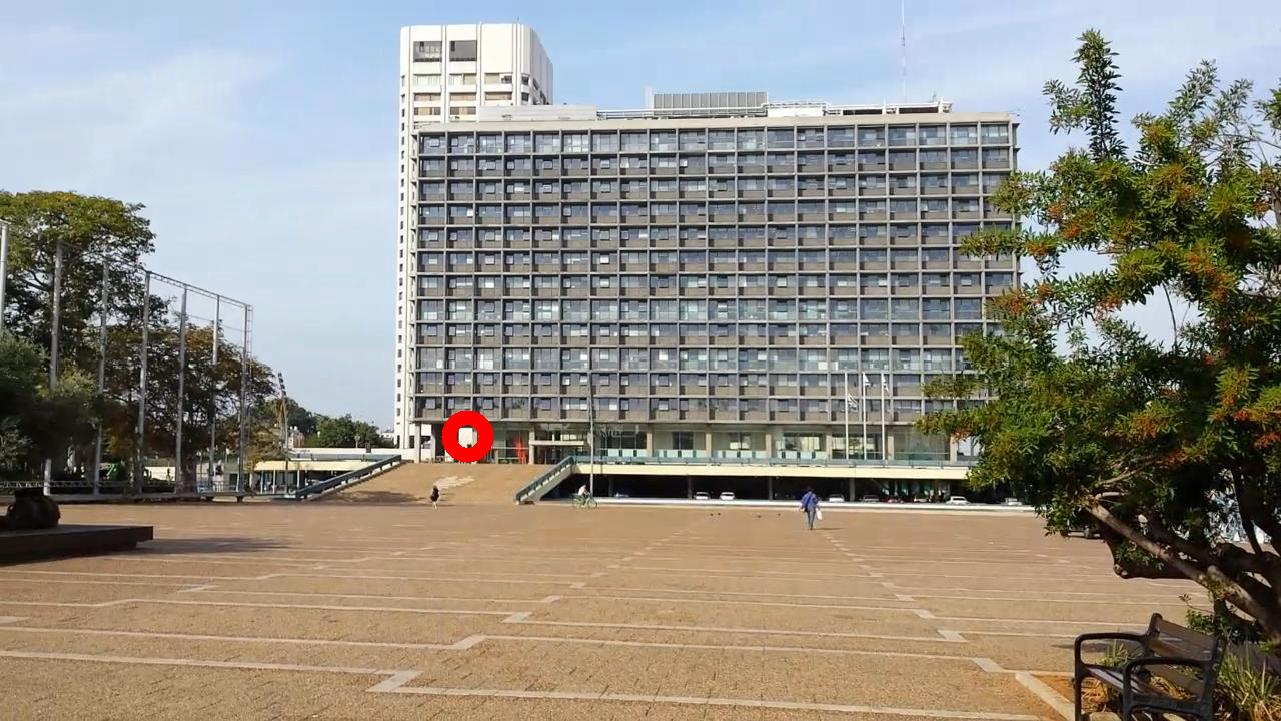} & \includegraphics[width=0.195\linewidth]{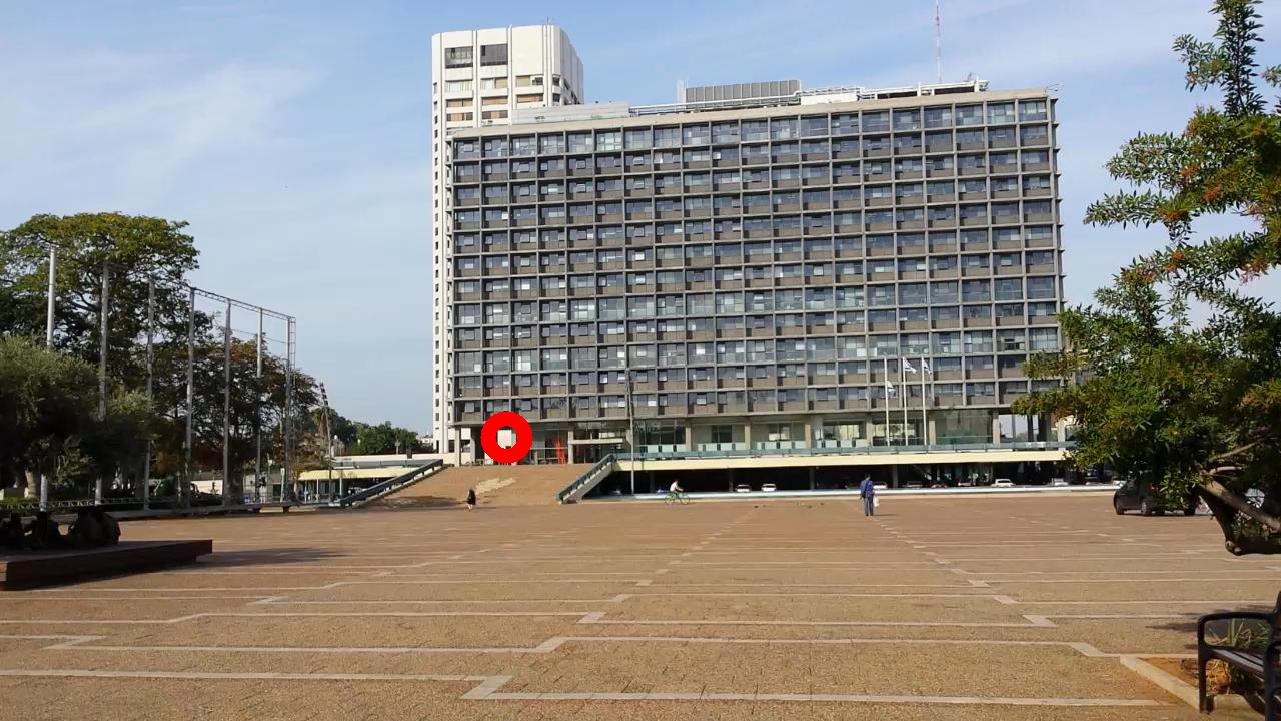} & \includegraphics[width=0.195\linewidth]{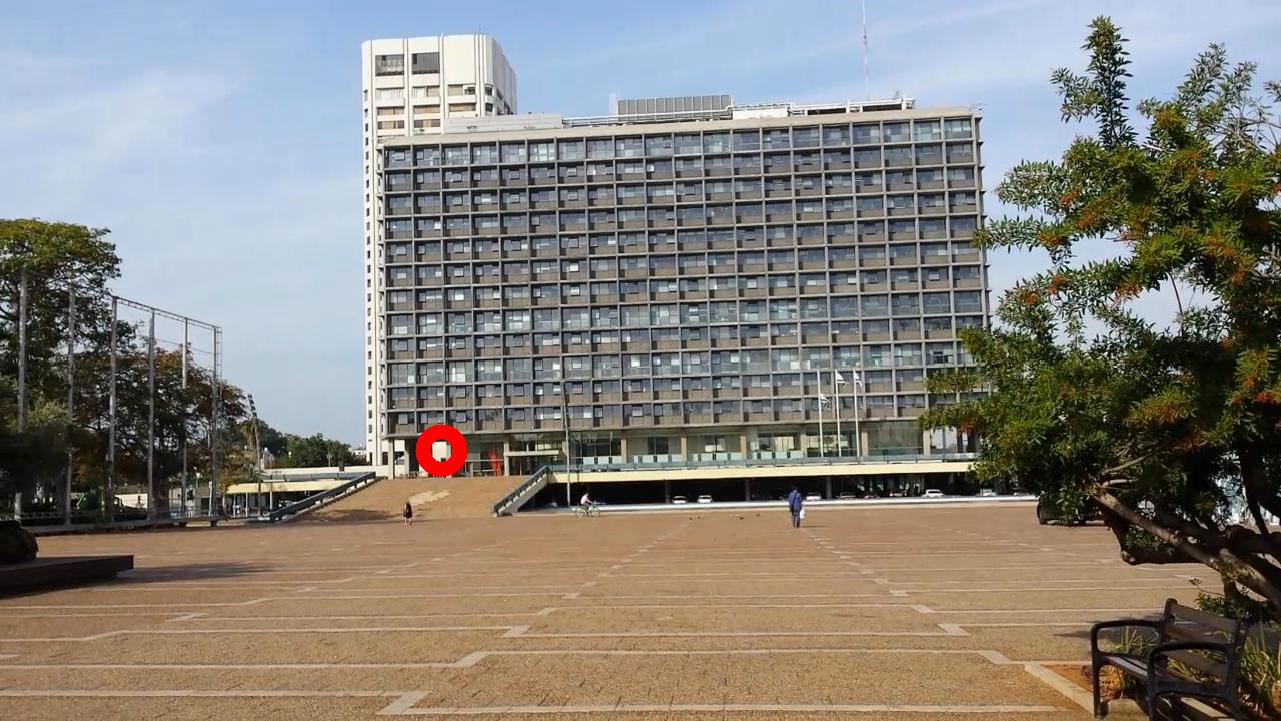} \\
			(a) & (b) & (c) & (d) & (e)
			\end{tabular}
		\caption{Camera guidance example with the supporting sets of images from dataset \emph{urban3}. Top row -- intermediate views and destination image; bottom row -- (a) and (b) are the support set of $I^0$, and (c)-(e) are the support set of $I^1$.}
\label{fig:figure2}
\end{figure*}




\renewcommand{\tabcolsep}{6pt}
{\small
\begin{table*}[t]
	\begin{center}
    		\begin{tabular}{ | c | c | c | c | c | c | c | }
    		\hline
				 & \textbf{\# images} & \textbf{Success rate} & \textbf{\# iterations} & \head{2.5cm}{\textbf{Cam.~ guidance time (min)}} & \head{2cm}{\textbf{Dictionary time (min)}} & \head{1.8cm}{\textbf{SFM time (min)}} \\ \hline
   		\textbf{\emph{urban1}} & 173 & 100\% & 2.6 & 3.2 & 3.7 & 215.9 \\ 
\hline
   		\textbf{\emph{urban2}} & 197 & 91.6\% & 2.42 & 1.7 & 3.8 & 289.2 \\ \hline
			\textbf{\emph{urban3}} & 235 & 83.3\% & 2 & 1.9 & 8.7 & 433.3 \\ \hline
			\textbf{\emph{park1}} & 90 & 91.6\% & 2.41 & 1.4 & 1.2 & 74.8 \\ \hline
			\textbf{\emph{park2}} & 141 & 91.6\% & 1.66 & 1.1 & 3.1 & 154.1 \\	\hline
			\textbf{\emph{park3}} & 206 & 91.6\% & 1.75 & 0.8 & 3.2 & 277.8 \\	\hline
    		\end{tabular}
	\end{center}
\caption{Camera guidance method: quantitative results}
\label{tab:table1}
\vspace{-0.3cm}
\end{table*}
}

\section{Implementation Details}
\label{sec:details}
Our method is run on a client-server configuration. The method without the interface (Sec.~\ref{sec:centralize}) was implemented in Matlab and run on a laptop (Toshiba Portege Z830-10D) as the server. The user images ($I^j$) were captured by a smartphone (client) and transferred via WiFi to the laptop. The interface was implemented for android smartphones using OpenCV4Android \cite{opencv4android}.
SIFT \cite{lowe2004distinctive} and SURF \cite{bay2008speeded} are too slow when computed on the smartphone for real time computation of the homographies that are used by the interface. Instead we use the Fast Retina Keypoint (FREAK) \cite{alahi2012freak}, which provides real-time performance.
\vspace{-0.3cm}
\paragraph{Parameter setting:}
The depth of recursion in the hierarchical K-means was set to $d=3$. Each layer consists of $K=\sqrt[d]{k}$ clusters, where $k=t_{k}w$ is the total number of clusters, $t_{k}$ is a clustering parameter, and $w$ is the number of SIFT features in all the images. The performance of rank aggregation degrades significantly when the number of clusters in the dictionary, $k$, is "too small", i.e., the feature correspondence precision is low. Since the feature correspondence recall is not as important, we sacrifice it for high precision by setting $t_k$ high (typically $t_k = 0.9$). In our experiments we restricted the number of supporting images, $|{\cal I}_T^0|$, to a maximum of $4$, which was shown to be sufficient for computing the intersection point.

\section{Experimental Results}
\label{sec:experiments}

To test our method, we assembled photos from six scenes: three scenes that were captured in an urban environment, \emph{urban1, urban2} and {\em urban3}, and three in public parks, \emph{park1, park2} and {\em park3}, each containing hundreds of images, captured by three different cameras (Samsung Galaxy S4, Samsung Galaxy Note, and Apple iPhone 4). The size of all images is $1280\times 720$.\footnote{These datasets will be publicly available upon paper acceptance.}
\vspace{-0.2cm}
\paragraph{Experiment 1 (our datasets):}
We tested our method on each of the datasets with 12 different pairs of initial and destination images. Examples of typical tests are shown in Fig.~\ref{fig:figure1}, Fig.~\ref{fig:figure2} and in the figure on the first page,
where the leftmost and rightmost images correspond to $I^0$ and $I_d$ respectively. In all of the examples the camera $C$ needs to be rotated to the right in order to view $P_{\scriptscriptstyle T}$. The images in the middle show the progression of our algorithm, where in each iteration the captured image, $I^i$, is "closer" to $I_d$. Moreover, the visual user interface is superimposed on the frames of $I^0$ and $I^1$. 
Examples of the support sets are shown in Fig.~\ref{fig:figure2} (bottom row).
Additional run examples are presented in the supplemental material.

A run is declared a success if $P_{\scriptscriptstyle T}$ is projected to the center region of $I^m$. Failure occurred in two scenarios. The first is when the camera is rotated to capture a new image that has no overlapping images in ${\cal I}$.
The algorithm can detect and alert the user about this case. This type of failure occurred rarely. The second is when the final image $I^m$ does not contain the projection of $P_{\scriptscriptstyle T}$ in its center region. This type of failure occured once in our tests.

The algorithm succeeds even when there is a large gap between $I^0$ and $I_d$, which requires the user to rotate the camera by over $100^\circ$. The assumption that the spatial orders of features are preserved in all images does not hold in our datasets due to moving objects (e.g., people), poles and trees, and feature correspondence errors (e.g., due to  repeated structures on the large building in Fig.~\ref{fig:figure2}). In addition, the computed fundamental matrices between pairs of images were sometimes inaccurate or missing. Our method copes with these challenges successfully. When only a single image in the support set is available,  the user rotates the camera to position the epipolar line in the frame. This often results in an additional intermediate view.

The method can also be applied when the point $P_{\scriptscriptstyle T}$ is occluded. See for example that the marked point (on the poles) in the support set of $I^0$ (Fig.~\ref{fig:figure3}(a)\&(b)) is occluded in $I^0$ by the tree. Hence, a direct computation of correspondence would fail in this case. Despite this, the point is successfully marked in $I^0$ by the intersection of the corresponding epipolar lines.  

To quantify the results the following measures are used: \begin{inparaenum}[(i)] \item success rate (success/total); \item mean number of intermediate views; 
\item time required for each run given the dictionary; \item time required for dictionary construction. \end{inparaenum} These results are summarized in Table~\ref{tab:table1}. The success rate of the algorithm is very high, and the average number of intermediate views in our experiments is $\sim$2. 

The overall running time of the algorithm consists of the offline dictionary construction step and the online camera guidance. 
The online camera guidance includes the reaction time of the user as well as the running time of the algorithm (SIFTs, homography, epipolar geometry and rank aggregation computation). The online components require about two minutes to run, which is reasonable as a proof of concept. We believe that this may be considerably improved using an optimized configuration, which will result in a real time application.

\paragraph{Experiment 2 (photo-tourism datasets):}
Although our assumption is that a dataset, $\setD$, is captured by photographers present at the scene close to the time at which the method is first used, we also tested our method with two publicly available datasets, Notre Dame and Trevi fountain \cite{snavely2008modeling}. Since most images in these datasets overlap, they were vertically cut in the middle to produce more challenging, non-overlapping images. We tested the first step of our method on randomly chosen pairs of $I^0$ and $I_d$. Additional steps require visiting these scenes. 
It seems that the points chosen by our method will result in rotating the camera towards $P_{\scriptscriptstyle T}$.
The first iteration results are fully presented in the supplemental material.


\renewcommand{\tabcolsep}{6pt}
{\small
\begin{table}[t]
	\begin{center}
    		\begin{tabular}{ | c | c | c | c | c | }
    		\hline
				 \textbf{\# im} & \head{1.2cm}{\textbf{SFM time (min)}} & \head{1.5cm}{\textbf{Cam. recovery rate}} & \head{1.1cm}{\textbf{Success rate (SFM)}} & \head{1.1cm}{\textbf{Success rate (ours)}} \\ \hline
					20 & 6.1 & 8~/~20 & 0/5 & 1/5  \\	\hline
					40 & 18.4 & 19.6~/~40 & 2/5 & 3/5 \\	\hline
					60 & 41.5 & 35.8~/~60 & 2/5 & 3/5  \\	\hline
					80 & 78.5 & 66.6~/~80 & 3/5 & 4/5 \\	\hline
					100 & 129.6 & 81.2~/~100 & 4/5 & 4/5 \\	\hline
					120 & 186.4 & 119.6~/~120 & 5/5 & 5/5 \\	\hline
    		\end{tabular}
	\end{center}
\caption{SFM results and running time for dataset \emph{park3}}
\label{tab:table2}
\vspace{-0.3cm}
\end{table}
}
\vspace{-0.2cm}
\paragraph{Comparison with SFM - running time:}
SFM may be used to project $P_{\scriptscriptstyle T}$ to $I^0$ (see Sec.~\ref{sec:method}). Then, our user interface may be used to center the projection of $P_{\scriptscriptstyle T}$ in the new captured image. But we wished to avoid using SFM due to its computational time, which is expected to be very high.
To quantify this claim, we compared the running time of our method with a state-of-the-art SFM method \cite{wu2011visualsfm,wu2013towards} based on "Bundler" \cite{snavely2008modeling} and parallelized using the GPU.
The running time of SFM for the datasets is presented in Table~\ref{tab:table1}. As expected, it is much slower than our method. In all datasets, our method is between 50 to 85 times faster than SFM, and it may take hours for the SFM to run. 
It can also be seen that as the number of images grows, the running time of SFM grows about quadratically.
\vspace{-0.2cm}
\paragraph{Comparison with SFM - \# of images:}
Since the number of images in the datasets was chosen somewhat arbitrarily, we tested the performance of the SFM and our method with different size subsets of images, randomly selected from the \emph{park3} dataset.
Two measures are used to quantify the performance. 

The first is the camera recovery ratio, $|C'|/|C|$, between the number of successfully recovered cameras, $C'$, and the total number of cameras, $C$. As the number of images grows, this ratio grows. Only above 120 images is it close to $1$; however, in this case, running the SFM takes hours. In scenes that were not captured beforehand, this duration is unacceptable. Table~\ref{tab:table2} presents the running time of the SFM method. Note that for each subset size (rows of Table~\ref{tab:table2}), 5 instances of random subsets were used, and the values in Table~\ref{tab:table2} are their average. 

The second measure is the success rate in five online runs of the camera guidance method (five runs per subset size). For each run, we use both the camera guidance with our full method, and the camera guidance based on directly computing the SFM (described in Sec.~\ref{sec:method}). The input to both methods, $I^0$ and $I_d$, vary from run to run. The results are comparable, with a slight advantage towards our method. Table~\ref{tab:table2} presents the results for the second measure. These results confirm the necessity of a faster alternative to SFM in the camera guidance problem. It is important to note that some of the $I_d$ that the SFM failed to recover, where successfully guided to by our method; thus, making our method applicable to "fill gaps" in photo collections used by SFM as input.
\vspace{-0.2cm}
\paragraph{Image based panoramas:}
To demonstrate that images in our datasets are generally not related by homography transformations, we used 10 pairs of overlapping images (from all datasets) that are separated by translation as input to a RANSAC procedure to estimate homography transformations. In most cases, there were no homography transformation with more than 4 inliers ("false" homographies). In these cases, any point besides the 4 inliers is transferred to a wrong location (usually very far) from the expected projection location. In two cases, a homography transformation was found with more than 30 inliers; however, it corresponds to planar surfaces in the images. In this case, any point not on the plane is transferred to a wrong location as well.

\section{Conclusions and Future Work}
\label{sec:conclusions}

We propose a new problem, called the camera guidance problem, whereby we wish to instruct a user to rotate his camera and capture an image such that it has overlapping FOV with a destination image. Our solution consists of two components. The first is to define the rotation and the second is to convey it to the user. 


Although SFM methods can be used to solve the camera guidance problem, we have shown that our method is much faster, with comparable performance. This is especially true in the common case, where dozens of images are required to obtain reasonable models of the scene and the cameras. 
We introduce the alternative SOFA scene representation and show that it can be efficiently computed using a rank aggregation approximation algorithm. It remains to be seen how the SOFA representation can be used for other tasks that require only the recovery of rough scene~geometry.




While it has been shown to be effective, our method is limited to specific settings, where the viewers are positioned on one side of the scene, e.g., a crowd in front of a stage. In addition, our method has been shown to tolerate a small number of moving objects and it is based on static regions. It would be interesting to study whether the moving objects can also be used for solving the camera guidance~problem.

{\small
\bibliographystyle{ieee}
\bibliography{cite}
}

\end{document}